\documentclass[runningheads]{llncs}

\usepackage{eccv}

\usepackage{eccvabbrv}

\usepackage{graphicx}
\usepackage{booktabs}
\usepackage{pifont}
\usepackage{subcaption}
\usepackage{wrapfig}
\usepackage{multirow}
\usepackage{multicol}
\usepackage{kotex}
\usepackage{mathtools}

\usepackage[accsupp]{axessibility}  

\newcommand{\cmark}{\textcolor{green}{\ding{51}}}%
\newcommand{\xmark}{\textcolor{red}{\ding{55}}}%
\usepackage{hyperref}

\usepackage{orcidlink}
\usepackage{booktabs,subcaption,graphicx}
\usepackage{xcolor,colortbl}

\begin{document}

\title{TTD: Text-Tag Self-Distillation Enhancing Image-Text Alignment in CLIP to Alleviate Single Tag Bias}

\titlerunning{TTD}

\author{Sanghyun Jo$^*$\inst{1} \and
Soohyun Ryu$^*$\inst{2} \and 
Sungyub Kim\inst{2} \and 
Eunho Yang\inst{2,3} \and
Kyungsu Kim$^\dagger$\inst{4}  
}

\authorrunning{S. Jo et al.}

\institute{OGQ, Seoul, Korea \and
Graduate School of AI, KAIST, Daejeon, Korea \and
AITRICS, Seoul, Korea\and
Massachusetts General Hospital and Harvard Medical School, Boston, MA, USA \\
\email{\{shjo.april, kskim.doc\}@gmail.com}, \email{\{rsoohyun, sungyub.kim, eunhoy\}@kaist.ac.kr}
}

\maketitle
\def\thefootnote{$*$}\footnotetext{Equal contribution}\def\thefootnote{\arabic{footnote}}
\def\thefootnote{$\dagger$}\footnotetext{Correspondence to}\def\thefootnote{\arabic{footnote}}

\begin{abstract}
We identify a critical bias in contemporary CLIP-based models, which we denote as \textit{single tag bias}. This bias manifests as a disproportionate focus on a singular tag (word) while neglecting other pertinent tags, stemming from CLIP's text embeddings that prioritize one specific tag in image-text relationships. When deconstructing text into individual tags, only one tag tends to have high relevancy with CLIP's image embedding, leading to biased tag relevancy. In this paper, we introduce a novel two-step fine-tuning approach, \textbf{T}ext-\textbf{T}ag Self-\textbf{D}istillation (TTD), to address this challenge.  TTD first extracts image-relevant tags from text based on their similarity to the nearest pixels then employs a self-distillation strategy to align combined masks with the text-derived mask. This approach ensures the unbiased image-text alignment of the CLIP-based models using only image-text pairs without necessitating additional supervision. Our technique demonstrates model-agnostic improvements in multi-tag classification and segmentation tasks, surpassing competing methods that rely on external resources. The code is available at \url{https://github.com/shjo-april/TTD}.

\end{abstract}

\begin{figure}[!ht]
    \centering
    \includegraphics[width=\linewidth]{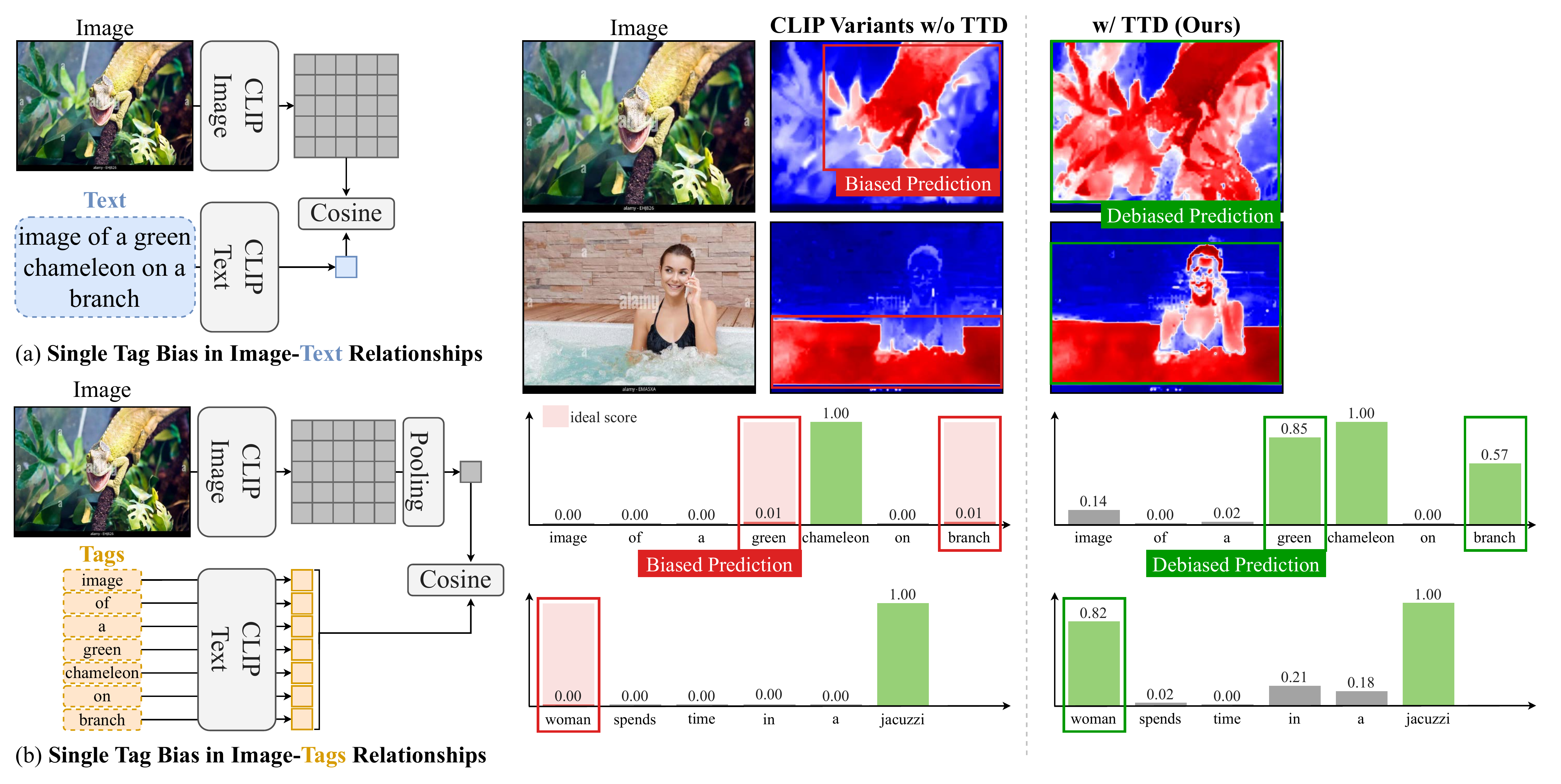}
\caption{\textbf{Single Tag Bias}. In existing image-text alignment research, \textit{single tag bias} is evident where the image and text embeddings tend to concentrate solely on a single tag. (a) \textbf{Single Tag Bias in Image-Text Relationships}: When examining the similarity map between the image (\emph{i.e.}, pixels) and text, it is evident that only a region of a single tag inside the red box is activated, disregarding other tags mentioned in the text. (b) \textbf{Single Tag Bias in Image-Tag Relationships}: Even when observing the similarity values between each tag in the text and the image, a high value is assigned only to the single tag, while other tags that describe the image (\emph{e.g.}, ``green'' and ``branch'') exhibit similar values to insignificant tags (\emph{e.g.}, ``of'' and ``a''). We use TCL~\cite{cha2023learning} for the analysis.}
\label{fig:single-tag-bias}
\end{figure}

\section{Introduction}
\label{sec:intro}
Research on fully supervised classification/segmentation is widely applied in the field of computer vision. Various weakly/semi-supervised and few-shot methods have emerged to reduce labeling costs. Recently, there has been active self-supervised research in image-text alignment following CLIP~\cite{radford2021learning}. Despite being trained solely on images and text (\emph{i.e.}, weak annotation), CLIP can be applied to downstream tasks that require stronger annotations, such as labels and masks.

A major existing limitation we discovered with CLIP-based models is what we term as \textit{single tag bias}, where they tend to overemphasize a particular context of image-text relationship. As observed in \cref{fig:single-tag-bias}(a), the similarity map between image and \textit{text} reveals activation only in the region related to a single tag/word (\emph{i.e.}, chameleon), while neglecting others (\emph{i.e.}, branch and green).  This observed encoder output issue, \textit{single tag bias}, indicates a flaw in CLIP's pre-trained text encoder, which over-focuses on one tag, ignoring others.

\begin{figure}[!t]
    \centering
    \includegraphics[width=\linewidth]{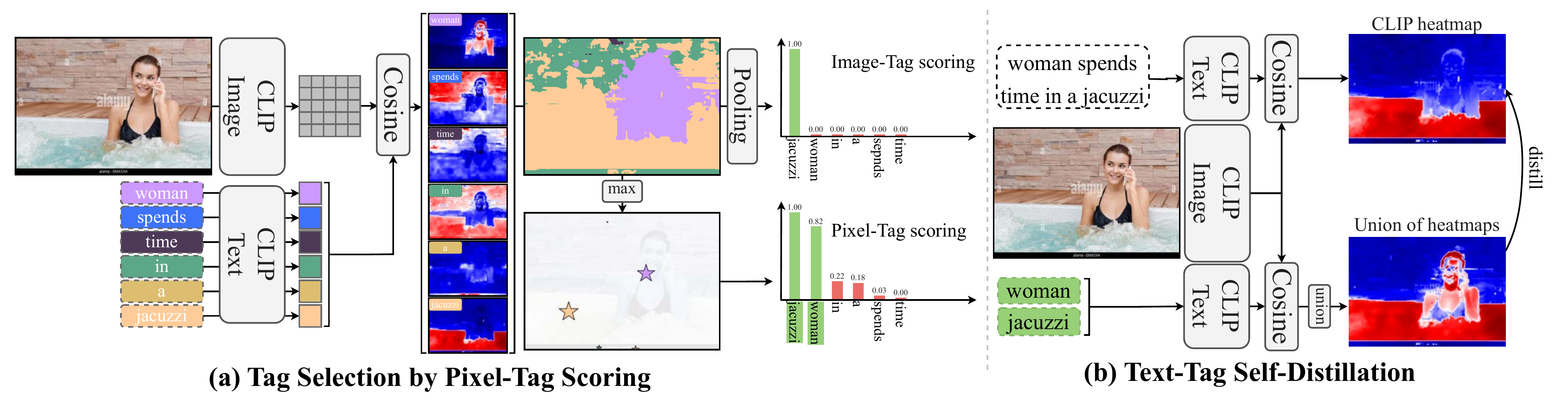}
\caption{\textbf{High-level Overview of Our Method}. (a) \textbf{Tag Selection by Pixel-Tag Scoring}: The global image embedding predominantly reflects information about a single tag, ``jacuzzi'' in this case, due to the overactivation of a single tag in pixel embeddings. The colored similarity map shows the most relevant tag for each pixel. (\emph{e.g.}, orange pixels have the highest similarity with ``jacuzzi''.) We minimize the impact of single tag bias observed in the image embedding by employing cosine similarity between the tag and its most correlated pixel as a score. (b) \textbf{Text-Tag Self-Distillation}: The similarity map between image and text demonstrates only a single tag, whereas it should represent all relevant tags in the image-text relationship. To alleviate the bias, we train the image-text map to align with the union of maps between the image and the pseudo-tags obtained in (a), thus enhancing the image-text alignment. }
\label{fig:overview}
\end{figure}

This paper aims to mitigate single tag bias in CLIP-based models using only image-text pairs, in line with CLIP's original design intent. Addressing this issue involves creating accurate foreground regions for every relevant tag, not just the biased one, to fine-tune the model to recognize their collective areas. It requires identifying all relevant tags in the text. However, the similarity between the image and each tag embedding, obtained by inputting each tag through CLIP's text encoder, shows persistent \textit{single tag bias} as seen in \cref{fig:single-tag-bias}(b). This underlines the struggle of standard CLIP-based tag scoring due to the image encoder's reduction of pixel data to a singular tag focus, blocking the recognition of other image-relevant tags. (\cref{fig:overview}(a) shows the pixels' overemphasis on one tag (\emph{i.e.}, ``jacuzzi''), skewing the image's global representation towards it.) Even when scoring based on the proportion of pixels similar to each tag within the image as in segmentation, it still fails to correctly extract meaningful tags from the text (see \cref{subsec:supp_tag}).

To address this, we shift from standard single image embedding to a new method focusing on the pixel nearest to each tag candidate. This is based on the finding that pixels closely correlated with a specific tag more accurately pinpoint image segments, as shown in \cref{fig:overview}(a). By utilizing the similarity between the nearest pixel and the tag as the measure for image-tag similarity, we mitigate bias and can accurately extract pseudo-tags that reflect the image-text relationship.

Based on extracted pseudo-tags, we create valid foreground masks for all image-relevant tags, removing unnecessary tag masks. This led to a new self-distillation-based CLIP fine-tuning method, assuming optimal image-text relationships encompass all pseudo-tag contexts (Fig. \ref{fig:overview}(b)). We create a pseudo-label, a union of the similarity maps between the image and all pseudo-tags, making CLIP learn this as the real image-text map. This process allows the model to recognize all relevant tag regions, not just one, improving CLIP's image-text alignment without additional annotations or models beyond CLIP.

In summary, our contributions are as follows:
\begin{itemize}
    \item Identified bias in pre-trained CLIP models focusing on a single tag in image-text relationships.
    \item Proposed new scoring and self-distillation for training CLIP models with image-related tags extracted from text automatically.
    \item Achieved superior multi-tag classification and segmentation by enhancing image-text alignment using solely image-text pairs without external models. 
    \item Provided new tag/mask annotations related to text inputs for future open-vocabulary research. 
\end{itemize}

\section{Related Work}
\label{sec:related_works}

CLIP~\cite{radford2021learning} stands out as the most prominent model for image-text alignment. It possesses a remarkable ability to comprehend various free-form texts, making it widely utilized in zero-shot/open-vocabulary scenarios. Consequently, various CLIP variants~\cite{cha2023learning, yi2023simple, zhou2022extract, liu2022open, ranasinghe2023perceptual} demonstrate comparable performance to models trained with task-specific annotations across many downstream tasks, such as classification and segmentation, even if they are trained with only image and text data. 

\begin{table}[!t]
\centering
\caption{\textbf{Comparison of Our Method with Related Approaches}.}
\label{tab:rebuttal1}
\resizebox{\textwidth}{!}{%
    \begin{tabular}{@{\extracolsep{1.5pt}}l|ccccccccc@{}}
        \toprule
        Properties & MaskCLIP & TCL & SimSeg & GroupViT & OVSeg & TagAlign & CoCu & CaSED & \textbf{Ours} \\
        \midrule
        (a) Select tags from text               & \xmark & \xmark & \xmark & \cmark & \cmark & \cmark & \cmark & \cmark & \cmark \\
        (b) Remove single tag bias    & \xmark & \xmark & \xmark & \xmark & \xmark & \xmark & \xmark & \xmark & \cmark \\
        (c) No external NLP models              & \cmark & \cmark & \cmark & \xmark & \xmark & \xmark & \xmark & \xmark & \cmark \\
        (d) Use image prior for tagging         & \xmark & \xmark & \xmark & \xmark & \xmark & \xmark & \cmark & \cmark & \cmark \\
        (e) Follow model-agnostic manner        & \xmark & \cmark & \xmark & \xmark & \cmark & \cmark & \xmark & \cmark & \cmark \\
        \bottomrule
    \end{tabular}%
}
\end{table}
\begin{figure}[!t]
    \centering
    \includegraphics[width=\linewidth]{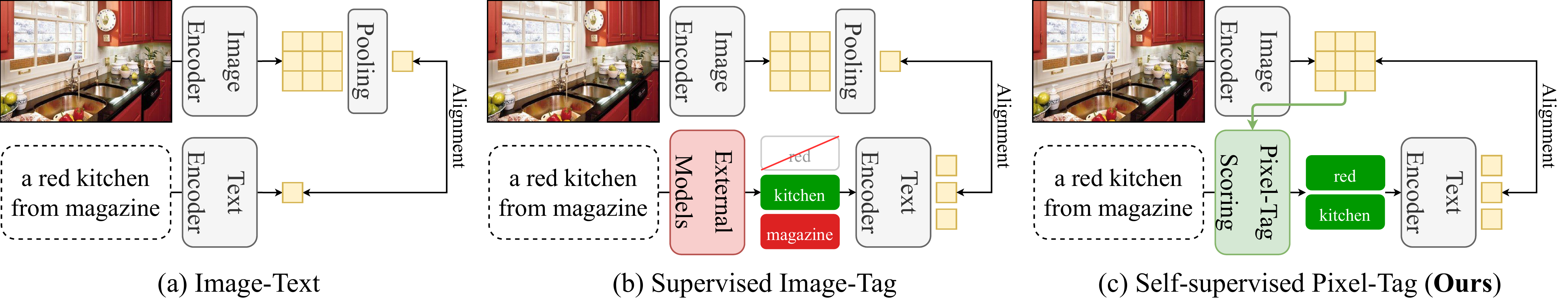}
\caption{\textbf{A Conceptual Comparison between Previous Approaches and Ours}. (a) Image-text alignment models, while useful for various open-vocabulary tasks, suffer from single tag bias. To alleviate this bias, extracting tags that reflect the image-text relationship from the text for training is crucial. (b) However, existing research often relies on external NLP models to extract tags without considering images, leading to issues: 1) Extracting image-irrelevant tags. 2) Overlooking image-relevant tags. (c) In contrast, we propose a tag selection method using only pixel information from images, eliminating the need for reliance on external models.}
\label{fig:alignment}
\end{figure}

\subsection{CLIP Variants with Pixel-Text Alignment}
CLIP faces challenges in downstream tasks requiring localization (\emph{e.g.}, segmentation and detection). Consequently, recent models~\cite{cha2023learning, yi2023simple, zhou2022extract, liu2022open, ranasinghe2023perceptual} have emerged that learn pixel-text alignment solely from image-text data without additional annotations. 
TCL~\cite{cha2023learning} generates a text-grounded mask that represents the relationship between an image and corresponding text, then aligns the text-grounded image and the text. SimSeg~\cite{yi2023simple} aims to align a sparse portion of the image and the text that only contains critical information. While these studies showcase impressive performance, they still exhibit single tag bias in image-text relationships as shown in \cref{tab:captioniou} as they implicitly learn alignment from the text. In contrast, our method extracts meaningful tags from the text and uses them for learning to alleviate the bias. 

\subsection{CLIP Variants with Tag Selection}  

\subsubsection{Utilization of External NLP models} 
Recent studies~\cite{liang2023open, mukhoti2023open, xu2022groupvit, xu2023learning, liu2023tagalign} have focused on extracting tags from text for learning, relying on external NLP models (\cref{fig:alignment}(b)). While some studies utilize NLP parsers~\cite{loper2002nltk, spacy2, akbik2019flair} to extract nouns, others leverage Large Language Models (LLM) like Vicuna~\cite{chiang2023vicuna}. However, these text-based tag selection techniques solely rely on text input and entirely disregard image information. This leads to two issues as shown in \cref{fig:qual_tag}: 1) Extracting image-irrelevant tags. 2) Overlooking image-relevant tags, particularly non-nouns. In contrast, our method avoids these issues by exclusively utilizing both image and text for tag selection.

\subsubsection{Additional Filtering using Image Information} Some models~\cite{xing2024rewrite, conti2024vocabulary} additionally consider the image to score tags. However, they utilize globally pooled image representation, suffering from single tag bias (\cref{fig:single-tag-bias}(b)). CoCu~\cite{xing2024rewrite} additionally introduces scoring based on retrieved image relevancy, but it depends on the images retrieved, where tag-relevant images lead to a low score and vice versa, and requires high computational cost (see \cref{subsec:supp_tag}). Also, both models rely on external NLP models and fail to address the omission of image-relevant tags.

Our proposed tag selection method scores tags based on the relationship between a tag and its most correlated pixel without employing external NLP models or data. This method mitigates single tag bias of image embedding and problems of external NLP models as shown in \cref{tab:ovc_tagacc}. Then, we present a text-tag self-distillation method to enhance pixel-tag alignment to address single tag bias in image-text relationships. As this approach uses only image and text data without additional annotations, it can be applied in a model-agnostic manner.

\begin{figure}[!t]
    \centering
    \begin{subfigure}{\textwidth}
        \includegraphics[width=\textwidth]{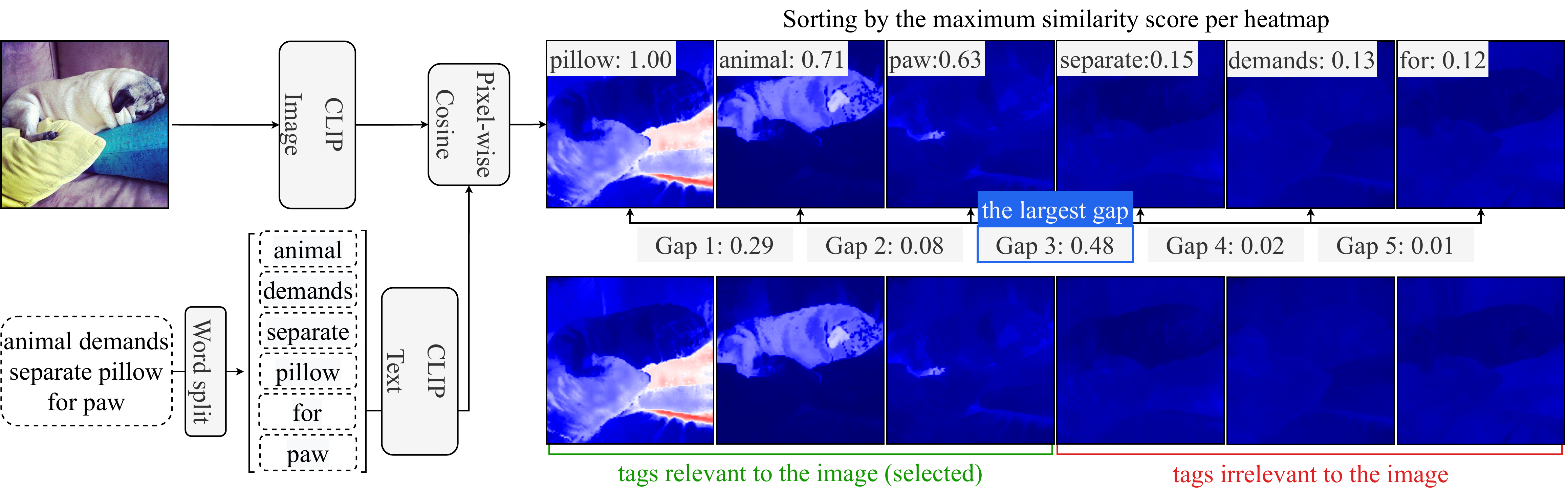}
        \caption{Tag Selection by Pixel-Tag Scoring.}
        \label{fig:tagging}
    \end{subfigure}
    \hfill
    \begin{subfigure}{\textwidth}
        \centering
        \includegraphics[width=0.9\textwidth]{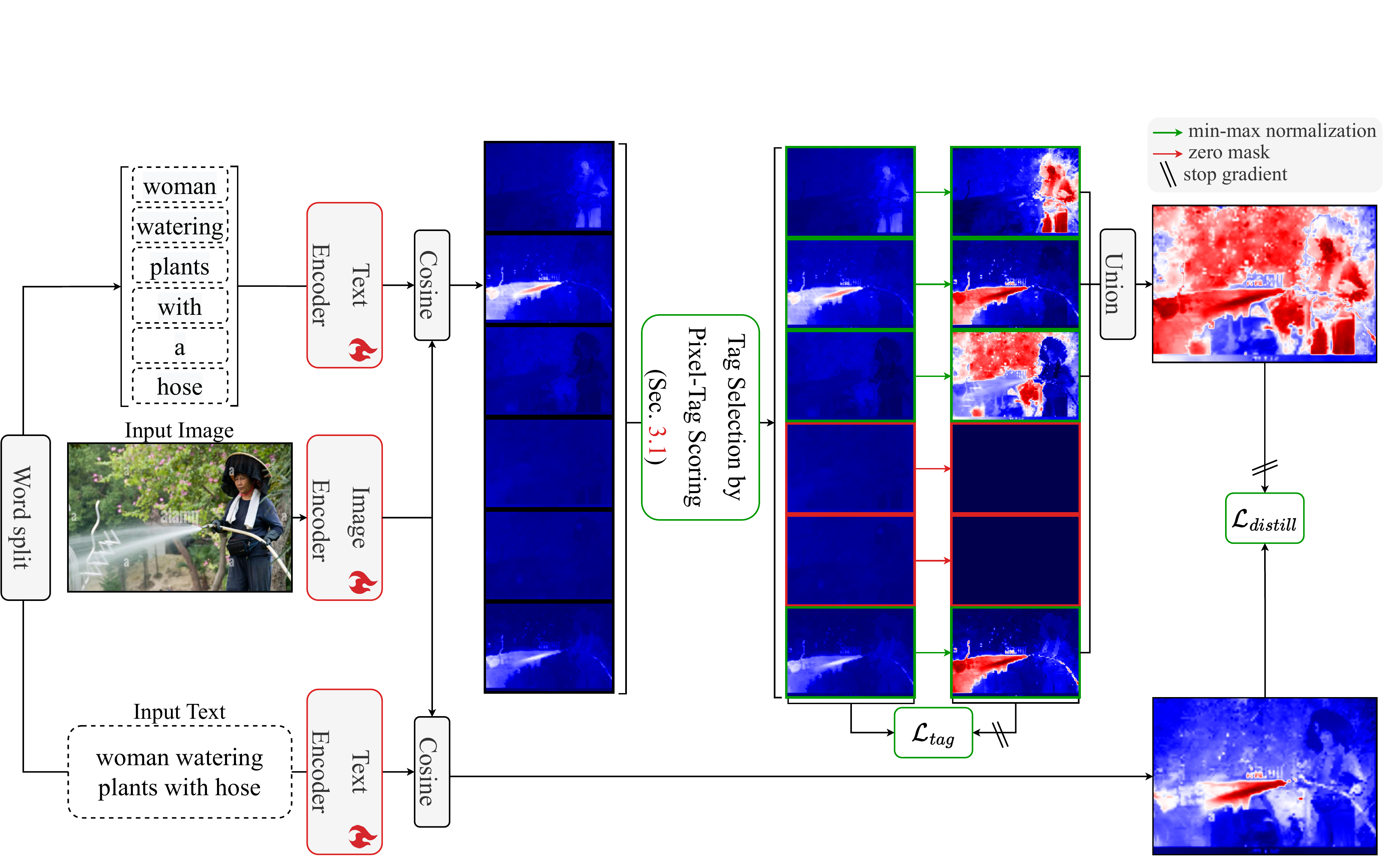}
        \caption{Method Overview.}
        \label{fig:overview_all}
    \end{subfigure}            
\caption{\textbf{Overall Framework of Our Method}. (a) \textbf{Tag Selection by Pixel-Tag Scoring}: Utilizing the similarity with the most correlated pixel as scoring, we identify a significant gap between tags related to the image and irrelevant ones. This gap guides us in selecting tags with the largest score gap as the threshold.
(b) \textbf{Method Overview}: Our fine-tuning approach begins by obtaining pseudo-tags through our tag selection process, which are then used to create an ideal heatmap representing the image-text relationship. We address single tag bias by distilling information between the ideal and actual heatmaps. Additionally, we integrate an auxiliary loss to learn the pixel-tag relationship. Throughout this process, both the image and text encoders are trained.
}
\label{fig:method-overview}
\end{figure}

\section{Method}
\label{sec:method}

Our fine-tuning method that alleviates single tag bias consists of two stages: Tag Selection by Pixel-Tag Scoring (\cref{subsec:tagging}) and Text-Tag Self-Distillation (\cref{subsec:fine-tune}). The overall framework is illustrated in \cref{fig:method-overview}.

\subsection{Tag Selection by Pixel-Tag Scoring}
\label{subsec:tagging}

Given an image $I$ of size $H\times W$ and corresponding text $T$ composed of tags $ \mathcal{T}=\{t_1 \cdots t_n\}$, our objective is to extract tags, $\mathcal{T}_{gt} \subset \mathcal{T}$ that accurately represent the image-text relationship. (\emph{e.g.}, for the first image in \cref{fig:qual_tag}, $\mathcal{T}_{gt}=\{t_1, t_2, t_3, t_6, t_7\}$.) Importantly, our approach relies solely on the image $I$ and the text $T$, without any external datasets or models.

Let $\mathcal{E}_V, \mathcal{E}_T$ denote the image and text encoders, respectively, which provide pixel embeddings and global text embeddings: $\mathcal{E}_V(I)\in \mathbb{R}^{H\times W\times C},\mathcal{E}_T(T)\in \mathbb{R}^C$ where $C$ is the embedding dimension size of the encoder output and $H, W$ are height and width of the image. The global image embedding can be obtained by pooling the pixel embedding: $\mathrm{pool}(\mathcal{E}_V(I)) \in \mathbb{R}^C$. 

A simple approach involves using the similarity between the image and a tag as a score:
\begin{equation} \label{eq:s_image}
    s_{\mathrm{image}}(t_i) := \mathrm{sim}(\mathrm{pool}(\mathcal{E}_V(I)), \mathcal{E}_T(t_i)),
\end{equation}
where $\mathrm{sim}$ denotes similarity metric, and cosine similarity is used in our case. However, this scoring method is affected by single tag bias, as depicted in \cref{fig:single-tag-bias}(b). This bias originates from the tendency of most pixels to be biased towards a specific tag, as shown in \cref{fig:overview}(a), thereby overlooking other relevant tags.

To mitigate this issue, we propose a novel scoring approach that leverages the pixel embedding most correlated with a tag rather than relying on a global image embedding:
\begin{equation} \label{eq:s_pixel}
    s_{\mathrm{pixel}}^{\mathrm{ours}}(t_i) := \max_{1\leq h \leq H, 1 \leq w \leq W} \mathrm{sim}(\mathcal{E}_V(I)_{hw}, \mathcal{E}_T(t_i)).
\end{equation}
This approach ensures more accurate consideration of objects within the image since each pixel corresponds to its most similar object. Utilizing pixel embeddings from the stage before pooling to compute similarity scores increases computational overhead compared to a simple approach. However, it can be reused because the computed similarity map is used for distillation in the subsequent step (\cref{subsec:fine-tune}).

Subsequently, we notice a substantial score gap between tags associated with the image and those not. As illustrated in \cref{fig:method-overview}(a), we arrange all tags according to $ s_{\mathrm{pixel}}^{\mathrm{ours}}$ and compute the score difference between consecutive tags. We observed the significantly large score difference between meaningful tags and others at the boundary. Leveraging this observation, we extract pseudo-tags with high scores using the largest score gap as the boundary: $\mathcal{T}_{pred} \subset \mathcal{T}$. 

Our pixel-tag scoring ($s_{\mathrm{pixel}}^{\mathrm{ours}}$) is derived through a straightforward calculation similar to the simple approach ($s_{\mathrm{image}}$). However, it showcases outstanding tag selection performance and effectively addresses issues encountered by methods employing external NLP models as seen in \cref{tab:ovc_tagacc}. Furthermore, unlike $s_{\mathrm{image}}$, which relies on heuristic methods like selecting the top-$k$ tags or setting a constant threshold due to insignificant score differences beyond the major tag, our tag selection method automatically identifies appropriate tags.

\subsection{Text-Tag Self-Distillation}
\label{subsec:fine-tune}
In this section, we introduce a self-distillation technique designed to mitigate single tag bias arising from text embedding (\cref{fig:single-tag-bias}(a)). The pseudo-tags $\mathcal{T}_{pred}$ obtained from our tag selection method (\cref{subsec:tagging}) encapsulate the relationship between the image and the text, enabling the transfer of knowledge from pseudo-tags to the text. Specifically, we ensure alignment between the union of similarity maps of pseudo-tags and the image with that of the text and image:
\begin{equation} \label{eq:simmap}
    \mathrm{simmap}(x) := \mathrm{sim}(\mathcal{E}_V(I), \mathcal{E}_T(x)) \in \mathbb{R}^{H\times W},
\end{equation}
\begin{equation} \label{eq:L_distill}
    \mathcal{L}_{distill} = ||\mathrm{simmap}(T) - \bigcup_{t_i\in \mathcal{T}_{pred}} \mathrm{norm}(\mathrm{simmap}(t_i))||^2_2,
\end{equation}
where $\mathrm{simmap}(x)$ is an abbreviation of the similarity map between the image $I$ and a text $x$, $\bigcup$ is the pixel-wise (element-wise) maximum operation, and $\mathrm{norm}$ is min-max normalization. Here, we apply a stop-gradient operation on the union similarity map to propagate the gradients only through the text-based one. 

Additionally, we introduce an auxiliary loss function to reinforce the correlation between tags and images. Non-selected tags are deemed irrelevant to the image; thus, we drive their similarity map toward zero. Conversely, selected tags are trained to align with the discrete map to disregard unrelated image parts:
\begin{equation} 
    \begin{multlined}
        \mathcal{L}_{tag} = \sum_{t_i \in \mathcal{T}} D(t_i) \\ \mathrm{, where} \ D(t_i) = \begin{cases}
        ||\mathrm{simmap}(t_i) - \mathrm{norm}(\mathrm{simmap}(t_i))||^2_2 &\mathrm{if} \ t_i \in \mathcal{T}_{pred}\\
        ||\mathrm{simmap}(t_i)||^2_2 &\mathrm{otherwise}
        \end{cases}.
    \end{multlined}
    \label{eq:L_tag}
\end{equation}
We also apply a stop-gradient operation on the normalized masks.

Our final fine-tuning loss function is defined as follows:
\begin{equation} \label{eq:L_final}
    \mathcal{L} = \mathcal{L}_{distill} + \mathcal{L}_{tag}.
\end{equation}

Some may question whether it is more effective to view segment-level relationships using external models like SAM~\cite{kirillov2023segment} rather than examining pixel-level relationships. However, our method is unrestricted and versatile since we self-train the model utilizing only image and text information without additional supervision, such as ground-truth tag annotation or external models. We also provide supplementary experiments on advanced techniques utilizing external models in \cref{subsec:supp_sam}.

\definecolor{md}{rgb}{1.0,0.2,0.2}
\definecolor{ud}{rgb}{0.2,0.6,1.0}

\section{Experiments}
\label{sec:exp}

\subsubsection{Datasets and Implementation Details}
In all experiments, we train our model solely with image-text pairs without utilizing additional annotations like ground-truth tag information for the text. We demonstrate its flexibility by evaluating its performance on two distinct public models~\cite{zhou2022extract, cha2023learning}. Furthermore, to cut computational costs, we utilize the LoRA~\cite{hu2021lora} technique for all our training processes. We use the CC3M and CC12M datasets~\cite{sharma2018conceptual, changpinyo2021conceptual} for training, applying an additional sample filtering strategy to reduce the use of largely uncorrelated image-text pairs. 

To evaluate the performance of our tag selection method, we manually collect ground-truth tag information for all image-text pairs of CC3M validation dataset~\cite{sharma2018conceptual}. Also, to see how much the single tag bias has eased, we generate the tag's corresponding mask data for 680 image-text pairs. This is necessary because, to our knowledge, there is no existing dataset that provides a complete match of all ground-truth tags and their corresponding masks with the text. We will make this information available through code release. Details are in \cref{sec:supp_dataset}. 

Additionally, for assessing our method's effectiveness in open-vocabulary semantic segmentation using tag inputs, we use six established benchmarks: PASCAL VOC (VOC)~\cite{everingham2010pascal}, PASCAL Context (Context)~\cite{mottaghi2014role}, COCO-Stuff (Stuff), COCO-Object (Object)~\cite{caesar2018coco}, Cityscapes (City)~\cite{cordts2016cityscapes}, and ADE20k (ADE)~\cite{zhou2019semantic}. Further evaluations on segmentation performance with referring datasets, alongside conventional open-vocabulary classification tasks, and specifics of these assessments are detailed in \cref{subsec:supp_impl} and \ref{sec:supp_add_exps}.

\subsubsection{Evaluation Metrics}
For evaluating tag selection, we use metrics like precision, recall, accuracy, F1 score, and mean Average Precision (mAP) as established in existing studies~\cite{he2023open, li2023patchct}. Precision measures the fraction of accurately identified positive tags among all predicted as positive, and recall quantifies the fraction of accurately identified positive tags out of all true positive tags. Notably, since there is no predetermined label set, we adopt sample-wise mAP~\cite{kobayashi2023two} over the standard class-wise mAP.

To evaluate segmentation performance, we use mean Intersection over Union (mIoU). For our proposed benchmark, considering the assessment of a model's ability to capture the relationship between captions and images, we refer to this mIoU as CaptionIoU. Furthermore, given our binary mask annotation, we also report the mean false positive rate (mFPR) and mean false negative rate (mFNR) to provide a comprehensive assessment of segmentation quality.

\subsection{Main Results}
\label{subsec:exp_main}

\begin{figure}[t]
    \centering
    \includegraphics[width=0.9\linewidth]{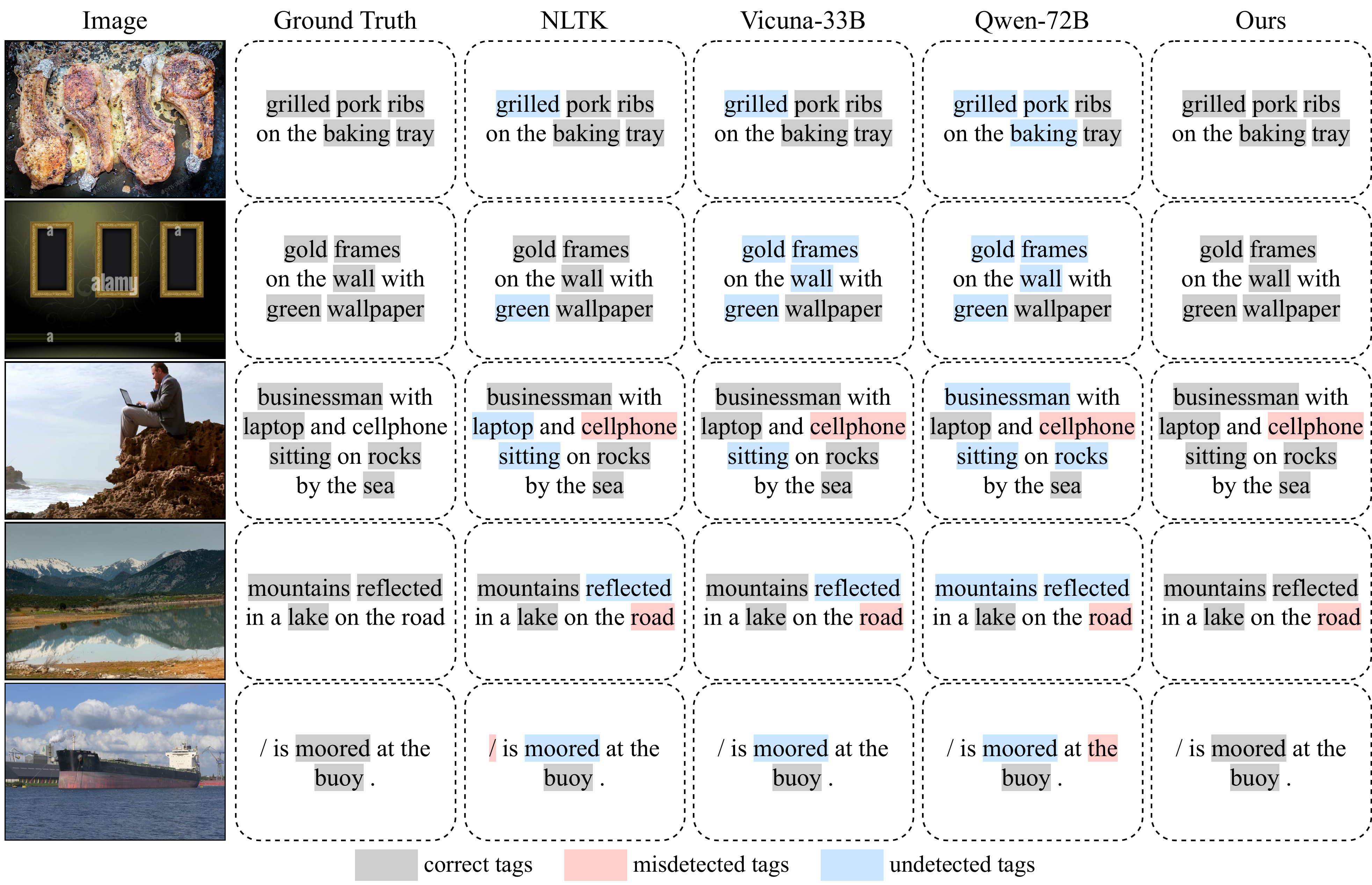}
\caption{\textbf{Qualitative Results of Tag Selection}. Tag selection using external models~\cite{loper2002nltk, chiang2023vicuna, bai2023qwen} suffers from two problems because they do not consider image information. {\textbf{1) Extracting Image-irrelevant Tags (\textcolor{md}{red})}: Select tags that do not correlate with the corresponding image content.} {\textbf{2) Overlooking Image-relevant Tags (\textcolor{ud}{blue})}: Cannot select tags related to the image, particularly non-nouns in the case of NLTK.}}
\label{fig:qual_tag}
\end{figure}
\definecolor{pink}{rgb}{0.96,0.82,0.82}

\setlength{\tabcolsep}{1.5pt}

\begin{table}[!t]
  \caption{\textbf{Multi-Tag Selection}. The best results are \textbf{bold} and the second best results are \underline{underlined}. P, precision; R, recall; F1, F1 score.}
  \label{tab:ovc_tagacc}
  \centering
  \begin{subtable}[t]{0.49\linewidth}
    \centering
    \caption{Comparison based on the use of NLP models.}
    \label{tab:ovc_tagacc_nlp}
    \resizebox{0.65\linewidth}{!}{%
      \begin{tabular}{@{}lcccc@{}}
        \toprule
        Method & P & R & F1 & Acc\\
        \midrule
        \midrule
        NLTK        & 59.8 & \textbf{83.7} & 69.8 & 79.6 \\
        Vicuna-7B   & 44.1 & 71.0 & 54.4 & 70.9 \\
        Vicuna-33B  & 52.7 & 70.7 & 60.4 & 75.9 \\
        Qwen-72B    & 69.3 & 56.2 & 62.1 & 80.9 \\
        \midrule
        $s_{\mathrm{pixel}}^{\mathrm{ours}}$ (\cref{eq:s_pixel})    & \underline{82.9} & 74.5 & \underline{78.5} & \underline{88.6} \\
        + TTD (Ours)     & \textbf{88.3} & \underline{78.0} & \textbf{82.8} & \textbf{91.0} \\
        \bottomrule
      \end{tabular}%
    }%
  \end{subtable}
  \hfill
  \begin{subtable}[t]{0.49\linewidth}
    \centering
    \caption{Comparison based on scoring methods.}
    \label{tab:ovc_tagacc_scoring}
    \resizebox{0.8\linewidth}{!}{%
      \begin{tabular}{@{}lccccc@{}}
        \toprule
        Scoring & P & R & F1 & Acc & mAP\\
        \midrule
        \midrule
        $s_{\mathrm{image}}$ (\cref{eq:s_image})        & \textbf{92.5} & 28.6 & 43.7 & 79.5 & 83.2 \\ 
        $s_{\mathrm{text}}$ (\cref{eq:s_text})        & 85.6 & 29.7 & 44.1 & 79.0 & 82.1 \\
        $s_{\mathrm{image}}+s_{\mathrm{text}}$         & 85.5 & 45.1 & 59.0 & 82.6 & 84.5 \\
        \midrule
        $s_{\mathrm{pixel}}^{\mathrm{ours}}$ (\cref{eq:s_pixel})        & 82.9 & \underline{74.5} & \underline{78.5} & \underline{88.6} & \underline{90.3} \\
        + TTD (Ours)        & \underline{88.3} & \textbf{78.0} & \textbf{82.8} & \textbf{91.0} & \textbf{93.7} \\
        \bottomrule
      \end{tabular}%
    }%
  \end{subtable}
\end{table}

\subsubsection{Multi-Tag Selection}
\label{subsubsec:exp_tagacc}
We assess the performance of our pixel-tag scoring-based method (Sec. \ref{subsec:tagging}) in the multi-tag selection task (\emph{i.e.}, the performance to identify all image-related tags given only the text without prior knowledge of the tags such as pre-defined label set). This evaluation is conducted in comparison to methods that use NLP models~\cite{loper2002nltk, chiang2023vicuna, bai2023qwen} and other basic scoring techniques. For scoring, we utilize image and text encoders from TCL~\cite{cha2023learning}, setting the threshold for other scoring approaches at 0.5.

As depicted in \cref{fig:qual_tag}, tag selection with external NLP models encounters two main issues: identifying tags not pertinent to the image and missing relevant tags due to ignoring visual information. As seen in the last example, NLTK selects irrelevant tags (\emph{e.g.}, `/') due to the overfit on pre-trained sentence structures. Nonetheless, our method largely extracts tags relevant to the image, addressing significant issues in conventional NLP models, especially identifying tags relevant to the image.

This superiority is further evident in \cref{tab:ovc_tagacc}. Higher precision indicates better filtering of image-irrelevant tags, while higher recall denotes improved extraction of image-relevant tags. As shown in \cref{tab:ovc_tagacc}(a), NLTK, the most widely used NLP parser for tag selection, exhibits high recall but at the expense of low precision, as it extracts all nouns without considering image information. Large Language Models (LLMs) show inferior performance compared to ours due to hallucination issues and lack of image consideration. In contrast, our method outperforms NLP models in both F1 score and accuracy with a significant margin, showing even better performance after fine-tuning. This indicates that our method effectively addresses the limitations of NLP models.

\cref{tab:ovc_tagacc}(b) presents the tag selection performance with different scoring methods. Following \cite{conti2024vocabulary}, we define a score that measures the similarity between the text and tag in addition to image-tag scoring ($s_{\mathrm{image}}$):
\begin{equation} \label{eq:s_text}
    s_{\mathrm{text}}(t_i) := \mathrm{sim}(\mathcal{E}_T(T), \mathcal{E}_T(t_i)).
\end{equation}

While image-tag scoring (\(s_{\mathrm{image}}\)) demonstrates high precision, it experiences very low recall due to the single tag bias, often leading to the selection of only one tag so if misses other image-related tags. Similarly, text-tag scoring (\(s_{\mathrm{text}}\)) and their combination show comparable issues. However, our approach significantly mitigates the single tag bias, showing high recall. Also, it exhibits high accuracy and mAP, which indicates the effectiveness of our scoring method. We further examine sensitivity to different thresholds and alternative scoring techniques that incorporate external data beyond the image-text pair in \cref{subsec:supp_tag}.

\subsubsection{Text-level Semantic Segmentation}
\label{subsubsec:exp_captioniou}
\begin{figure}[!t]
    \centering
    \includegraphics[width=0.95\linewidth]{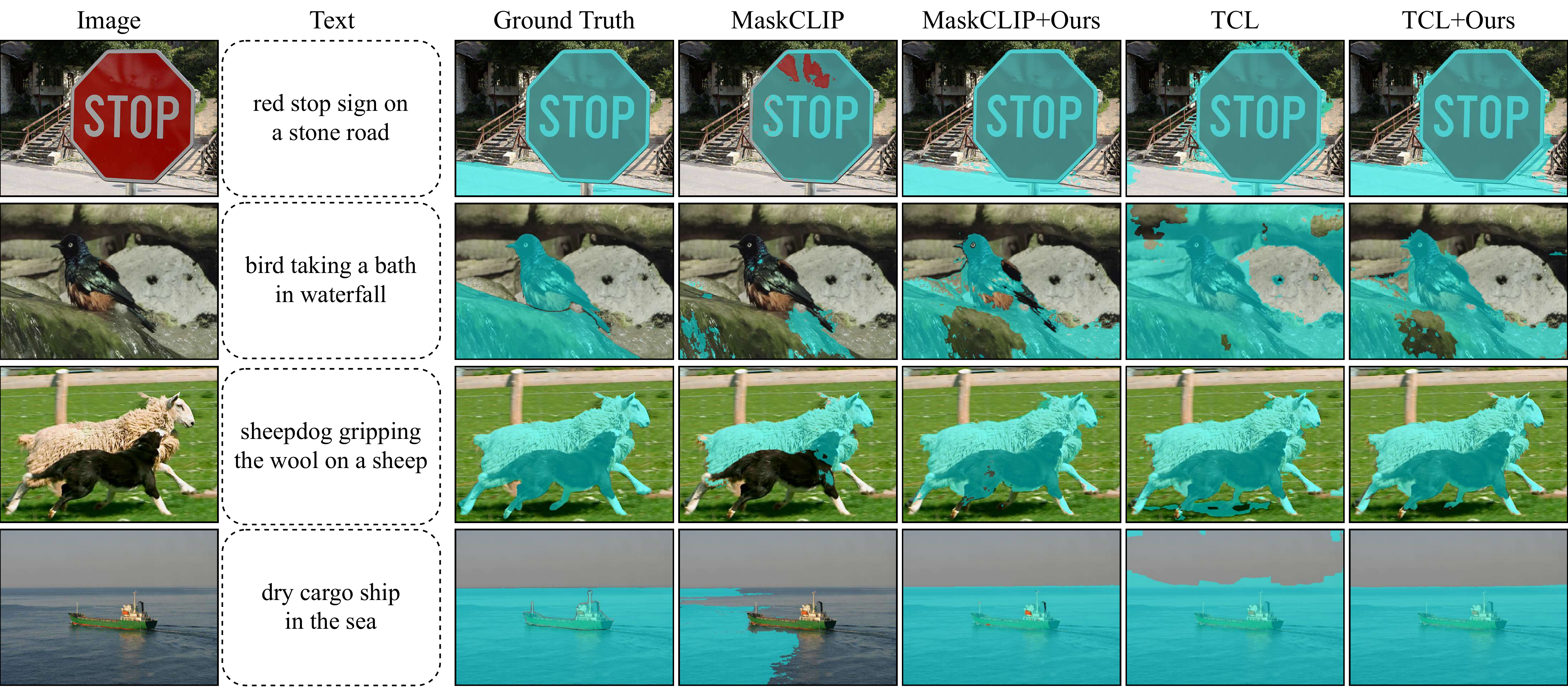}
\caption{\textbf{Qualitative Results on Text-level Segmentation}.}
\label{fig:qual_ciou}
\end{figure}

In this task, we evaluate text-level segmentation performance, where the text encoder of CLIP processes text input using existing CLIP-based models~\cite{cha2023learning, zhou2022extract} and followed by the application of our fine-tuning method, TTD. Specifically, this task evaluates whether the model accurately understands the relationship between the image and the given text. As depicted in \cref{fig:qual_ciou}, traditional models show a single tag bias, mainly capturing the spatial region of only one tag. TCL often generates masks that extend beyond the relevant context. Yet, after fine-tuning with our method, the single tag bias reduces, and the fine-tuned models accurately activate all regions associated with image-related tags in the respective text.

The effectiveness of TTD is further illustrated in \cref{tab:captioniou}, where its integration enhances the models' comprehension of the text-image relationship, leading to improved CaptionIoU scores (\emph{i.e.}, +7.2\% on average). Notably, MaskCLIP exhibits a significant decrease in the mFNR score (\emph{i.e.}, 0.169) post-fine-tuning, indicating its improved understanding beyond a single tag. Additionally, TCL demonstrates a noteworthy decrease in mFPR. It means that after fine-tuning, the model can exclude portions unrelated to the text, indicating that our fine-tuning method enhances model alignment as well.

\definecolor{pink}{rgb}{0.96,0.82,0.82}
\setlength{\tabcolsep}{1.5pt}
\begin{table}[!t]
  \caption{\textbf{Text-level Semantic Segmentation}. }
  \label{tab:captioniou}
  \centering
  \resizebox{0.4\textwidth}{!}{
    \begin{tabular}{@{}lccc@{}}
      \toprule
      Method & CaptionIoU(\%) & mFPR & mFNR \\
      \midrule
      \midrule
      MaskCLIP~\cite{zhou2022extract} & 41.0 & 0.179 & 0.411 \\
      \rowcolor{pink}
      + TTD (Ours) & 50.2 (+9.2) & 0.256 & 0.242 \\
      TCL~\cite{cha2023learning} & 60.4 & 0.199 & 0.198 \\
      \rowcolor{pink}
      + TTD (Ours) & 65.5 (+5.1) & 0.163 & 0.182 \\
      \bottomrule
    \end{tabular}%
  }
\end{table}
\vspace{-0.5cm}

\subsection{Results on Open-Vocabulary Semantic Segmentation}
\label{subsec:exp_ovs}

\begin{figure}[!t]
    \centering
    \includegraphics[width=0.8\linewidth]{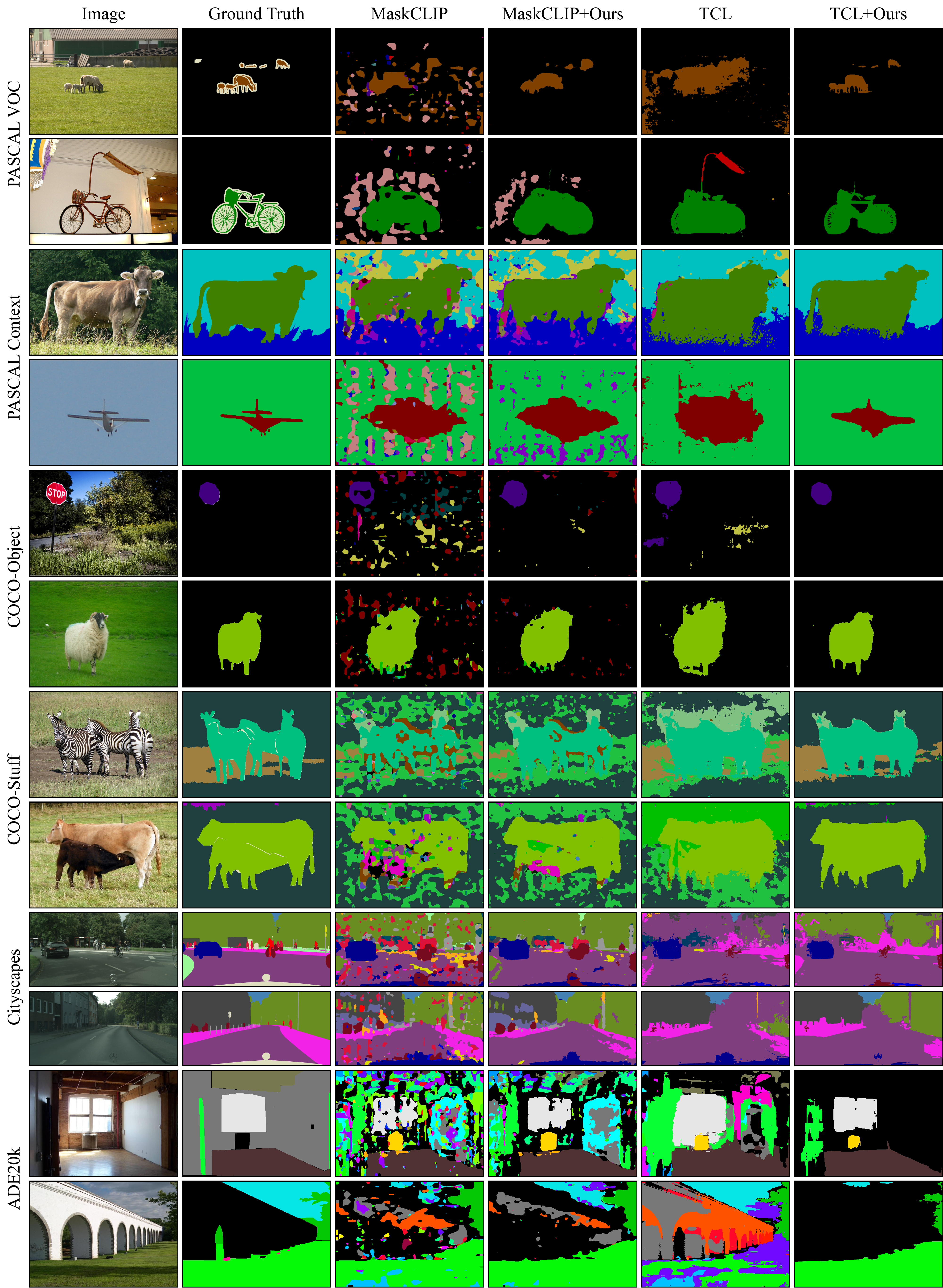}
\caption{\textbf{Qualitative Results on Tag-level Segmentation}.}
\label{fig:qual_ovs}
\end{figure}

In the previous task, we evaluated the classification and segmentation performance of TTD-trained CLIP using text input, measuring how well the model understands the relationship between the image and the text. Subsequently, we conducted experiments on standard open-vocabulary semantic segmentation where the text encoder was provided with tag input. Note that our TTD training did not utilize actual tag annotation from the text. In \cref{fig:qual_ovs}, we display segmentation results with and without TTD. Our approach effectively detects the presence of objects in the image, unlike conventional models that identify objects not actually present in the image.

\cref{tab:ovs} contrasts the tag-level segmentation performance of studies employing only image-text pairs without additional annotations or external models akin to ours for fair comparison. For this reason, models~\cite{liang2023open, ding2022decoupling, luo2023segclip, guo2023mvp} that utilize additional annotations or external models (\emph{e.g.}, mask proposal networks) are also excluded. Additionally, we compared models that utilize NLP-based models($\mathcal{L}$ in the table) for tag extraction. Implementing our fine-tuning technique results in an average increase in mIoUs of 8.5\% for MaskCLIP and 3.5\% for TCL. Furthermore, TCL enhanced with our method outperforms existing models on benchmarks such as Pascal VOC, Pascal Context, and COCO-Object. This indicates that our approach successfully augments model alignment in addition to alleviating single tag bias.

At this point, TagAlign~\cite{liu2023tagalign} achieved better performance on datasets that have at least 150 classes such as COCO-Stuff/ADE20K than our model. The reason for this is that TagAlign utilized multiple text information for a single image, whereas we opted for simplicity by using only a single text corresponding to each image. Therefore, the performance difference arises from the relatively insufficient positive/negative tag information in our method, indicating the potential value in further improving our fine-tuning method.

\definecolor{pink}{rgb}{0.96,0.82,0.82}

\setlength{\tabcolsep}{1.5pt}
\begin{table}[!tb]
  \caption{\textbf{Open-Vocabulary Semantic Segmentation}. The methods are all trained only with image and text data, without additional annotations or external models. We use ViT-B/16 as the backbone for all methods. $\mathcal{L}$, external language models.}
  \label{tab:ovs}
  \centering
  \resizebox{\textwidth}{!}{
  \begin{tabular}{@{}lccccccccc@{}}
    \toprule
    Method & Train datasets & $\mathcal{L}$ & VOC & Context & Object & Stuff & City & ADE & Avg. \\
    \midrule
    \midrule
    GroupViT~\cite{xu2022groupvit}    & CC12M+YFCC & \cmark & 51.1 & 19.0 & 27.9 & 15.4 & 11.6 & 9.4 & 22.4 \\
    ViewCo~\cite{ren2023viewco}          & CC12M+YFCC & \cmark & 52.4 & 23.0 & 23.5 & - & - & - & - \\
    CoCu~\cite{xing2024rewrite}        & CC3M+CC12M+COCO & \cmark & 51.4 & 23.6 & 22.7 & 15.2 & 22.1 & 12.3 & 24.6 \\
    OVSegmentor~\cite{xu2023learning}  & CC4M~\cite{xu2023learning} & \cmark & 53.8 & 20.4 & 25.1 & - & - & - & - \\
    TagAlign~\cite{liu2023tagalign}    & CC12M & \cmark & 53.9 & 33.5 & 33.3 & \textbf{25.3} & \underline{27.5} & \textbf{17.3} & \underline{31.8} \\
    ReCo~\cite{shin2022reco}          & ImageNet1K & \xmark & 25.1 & 19.9 & 15.7 & 14.8 & 21.1 & 11.2 & 18.0 \\
    ZeroSeg~\cite{chen2023exploring}          & ImageNet1K & \xmark & 40.8 & 20.4 & 20.2 & - & - & - & - \\
    ViL-Seg~\cite{liu2022open}          & CC12M & \xmark & 37.3 & 18.9 & 18.1 & - & - & - & - \\
    SimSeg~\cite{yi2023simple}          & CC3M+CC12M & \xmark & \underline{57.4} & 26.2 & 29.7 & - & - & - & - \\
    SegCLIP~\cite{luo2023segclip}          & CC12M+COCO & \xmark & 52.6 & 24.7 & 26.5 & 16.1 & 11.2 & 8.8 & 23.3 \\
    \midrule
    MaskCLIP~\cite{zhou2022extract}    & - & \xmark & 29.3 & 21.1 & 15.5 & 14.7 & 21.6 & 10.4 & 19.0 \\
    \rowcolor{pink}
    + TTD (Ours)              & CC3M+CC12M & \xmark & 43.1 (+13.8) & 31.0 (+9.9) & 26.5 (+11.0) & 19.4 (+4.7) & \textbf{32.0} (+10.4) & 12.7 (+2.3) & 27.5 (+8.5) \\
    TCL~\cite{cha2023learning}         & CC3M+CC12M & \xmark & 55.0 & \underline{33.8} & 31.6 & 22.4 & 24.0 & 15.6 & 30.4 \\
    \rowcolor{pink}
    + TTD (Ours)              & CC3M+CC12M & \xmark & \textbf{61.1} (+6.1) & \textbf{37.4} (+3.6) & \textbf{37.4} (+5.8) & \underline{23.7} (+1.3) & 27.0 (+3.0) & \underline{17.0} (+1.4) & \textbf{33.9} (+3.5) \\
  \bottomrule
  \end{tabular}
  }
\end{table}

\subsection{Ablation}
\label{subsec:exp_ab}
\setlength{\tabcolsep}{1.5pt}


\begin{table}[tb]
  \centering
  \caption{\textbf{Ablation Studies}.}
  \label{tab:ab}
  \begin{subtable}[t]{0.59\linewidth}
    \setlength{\tabcolsep}{1.5pt}
    \centering
    \caption{Effect of Loss Terms.}
    \label{tab:ab_loss}
    \resizebox{0.85\linewidth}{!}{%
      \begin{tabular}{@{}cccc@{}}
        \toprule
        $\mathcal{L}_{distill}$ (\cref{eq:L_distill}) & $\mathcal{L}_{tag}  (\cref{eq:L_tag})$ & CaptionIoU & mIoU\\
        \midrule
        \midrule
        \xmark & \xmark & 60.4 & 55.0 \\
        \xmark & \cmark & 60.7 (+0.3) & 58.5 (+3.5) \\
        \cmark & \xmark & 63.6 (+3.2) & 60.8 (+5.8) \\
        \cmark & \cmark & \textbf{65.5} (+5.1) & \textbf{61.1} (+6.1) \\
        \bottomrule
      \end{tabular}%
    }%
  \end{subtable}
  \hfill
  \begin{subtable}[t]{0.39\linewidth}
    \centering
    \caption{Effect of Tagging Method.}
    \label{tab:ab_tagging}
    \resizebox{0.75\linewidth}{!}{%
      \begin{tabular}{@{}lcc@{}}
        \toprule
        Method & CaptionIoU & mIoU\\
        \midrule
        \midrule
        Baseline~\cite{cha2023learning} & 60.4 & 55.0 \\
        \midrule
        NLTK~\cite{loper2002nltk} & 61.8 &  56.5 \\
        $s_{\mathrm{image}}$ (\cref{eq:s_image}) & 56.3 & 52.5 \\
        $s_{\mathrm{pixel}}^{\mathrm{ours}}$ (\cref{eq:s_pixel}) & \textbf{65.5} &  \textbf{61.1} \\
        \bottomrule
      \end{tabular}%
    }%
  \end{subtable}
  \hfill
  
\end{table}

In this section, we conducted ablation experiments to evaluate the impact of the techniques employed in our fine-tuning method. The assessment was performed using TCL~\cite{cha2023learning} on our text-level, and Pascal VOC~\cite{everingham2010pascal} for tag-level segmentation.

\subsubsection{Effect of Fine-tuning Losses} Our method incorporates two loss terms, $\mathcal{L}_{distill}$ and $\mathcal{L}_{tag}$, for fine-tuning. $\mathcal{L}_{distill}$ aims to mitigate single tag bias, while $\mathcal{L}_{tag}$ aims to enhance the pixel-tag correlation. When solely utilizing $\mathcal{L}_{tag}$, there was a marginal performance enhancement observed during the min-max normalization process for each mask, as depicted in \cref{tab:ab}(a). However, the inclusion of $\mathcal{L}_{distill}$ significantly alleviates single tag bias, leading to an improvement in both CaptionIoU and mIoU (\emph{i.e.}, +3.2\% and +5.8\%, respectively). The best performance is achieved when both losses are combined, learning each tag mask and union mask simultaneously.

\subsubsection{Effect of Tag Selection Method}
\label{subsubsec:exp_ab_tagging}
In our approach, tag selection is conducted via pixel-tag scoring (\cref{eq:s_pixel}) to extract pseudo-tags representing the image-text relationship. Alternatively, we examine NLTK and simple image-tag scoring (\cref{eq:s_image}) for tag selection, as shown in \cref{tab:ab}(b). When employing NLTK to extract nouns from texts, due to the aforementioned issues (extraction of image-irrelevant tags and omission of image-relevant tags), the model struggles to generate the ideal text-image similarity map (\emph{i.e.}, the union of pseudo-tag driven maps). Consequently, the generated union map may include the activation of unrelated tags or omit activations necessary for the image-tag relationship, resulting in minor improvement. On the other hand, using image-tag scoring tends to extract a single tag, exacerbating single tag bias and resulting in a decline in both CaptionIoU and mIoU. In contrast, our tag selection method exhibits significantly enhanced performance in both tasks, indicating that the pseudo-tags extracted using our method effectively capture the image-text relationship.

\section{Conclusion}
\label{sec:conclusion}
Our research tackles the disproportionate alignment between images and text inherent in existing CLIP-based models, which we term as single tag bias. To address this issue, we introduce a novel fine-tuning method TTD that uses only image-text pairs without additional annotations or external models. We identify all ground-truth tags related to the image within the text and then generate an ideal image-text map that reflects all related tags for self-distillation. Our method effectively mitigates single tag bias, improves overall image-text alignment, and enhances performance in both image segmentation and multi-tag classification within the text.
Our approach, following the TTD framework, enables the automatic generation of image-relevant tags, corresponding masks, and text masks, without reliance on external models beyond CLIP. This positions our method as a valuable tool for precise auto-labeling, with significant potential across various datasets consisting solely of images and texts.

\clearpage  

\appendix
\clearpage

\section{Automatic Labeling via TTD}
\begin{table}[!t]
\centering
\caption{\textbf{Comparison of Our Method with Related Approaches}.}
\label{tab:supp_rel}
\resizebox{\textwidth}{!}{%
    \begin{tabular}{@{\extracolsep{1.5pt}}l|ccccccccc@{}}
        \toprule
        Properties & CoCu & OVSeg & TagAlign & CaSED & GroupViT & MaskCLIP & SimSeg & TCL & \textbf{Ours} \\
        \midrule
        \textbf{Automatic Annotations} & & & & & & & & & \\
        (a) Generate tags from text               & \cmark & \cmark & \cmark & \cmark & \cmark & \xmark & \xmark & \xmark & \cmark \\
        (b) Generate tag-level mask & \xmark & \cmark & \xmark & \xmark & \cmark & \xmark & \xmark & \xmark & \cmark \\
        (c) Generate text-level mask & \xmark & \xmark & \xmark & \xmark & \xmark & \xmark & \xmark & \xmark & \cmark \\
        \midrule
        
        \textbf{Methodology} & & & & & & & & & \\
        (d) Alleviate single tag bias    & \xmark & \xmark & \xmark & \xmark & \xmark & \xmark & \xmark & \xmark & \cmark \\
        (e) No use of external models              & \xmark & \xmark & \xmark & \xmark & \xmark & \cmark & \cmark & \cmark & \cmark \\
        (f) Learning pixel-tag alignment    & \xmark & \xmark & \xmark & \xmark & \cmark & \xmark & \xmark & \xmark & \cmark \\
        (g) Follow model-agnostic manner        & \xmark & \cmark & \cmark & \cmark & \xmark & \xmark & \xmark & \cmark & \cmark \\
        \bottomrule
    \end{tabular}%
}
\end{table}

\begin{figure}[!t]
    \centering
    \includegraphics[width=\linewidth]{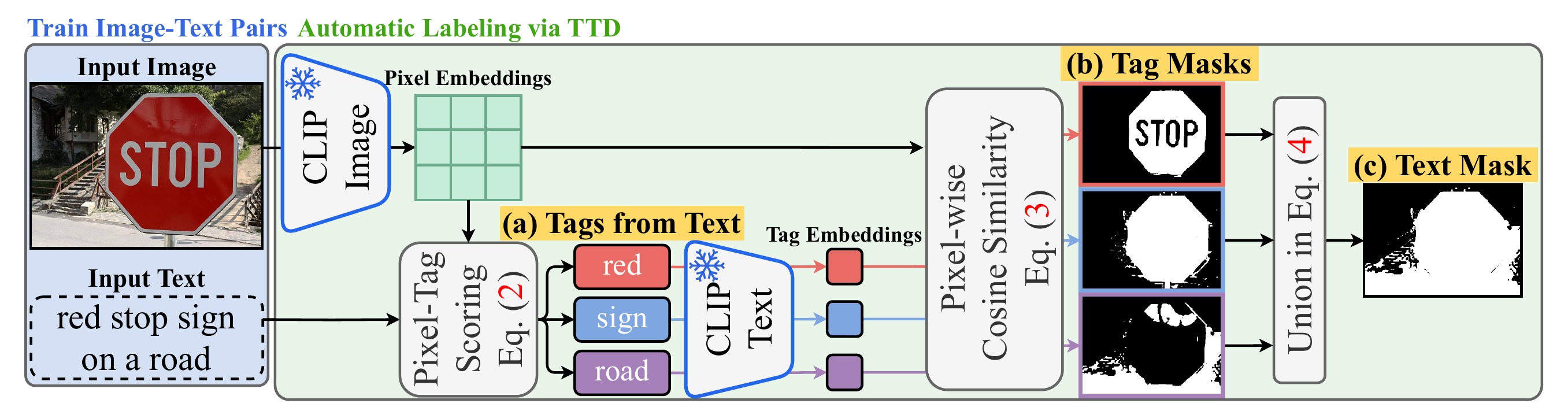}
\caption{\textbf{Automatic Labeling via TTD}. The proposed TTD framework enables the automatic generation of (a) relevant tags from the input text, (b) corresponding masks for each tag, and (c) a composite/debiased mask representing the entire text.}
\label{fig:alignment_ann}
\end{figure}

\setlength{\tabcolsep}{1.5pt}
\begin{table}[!t]
  \caption{\textbf{Performance of Automatic Labeling on the Train Set}. Training CLIP variants (\emph{e.g.}, MaskCLIP/TCL) with our TTD improves (a) Multi-Tag Selection and (b) Text-level Semantic Segmentation. We assess performance based on the manually collected tag and mask annotations of the train set (CC3M and CC12M datasets) using the same metrics as Tabs. \ref{tab:ovc_tagacc} and \ref{tab:captioniou}.}
  \label{tab:train_labeling}
  \centering
  \begin{subtable}[t]{0.535\linewidth}
    \centering
    \caption{Multi-Tag Selection.}
    \label{tab:rebuttal_tags}
    \resizebox{\linewidth}{!}{%
      \begin{tabular}{@{}lcccc@{}}
        \toprule
        Method & Precision & Recall & F1 & Accuracy \\
        \midrule
        \midrule
        MaskCLIP                  & 46.9 & 67.7 & 55.4 & 58.1  \\
        \rowcolor[HTML]{F4CCCC}
        + TTD                     & 62.5 & 78.2 & \textbf{69.5} & \textbf{73.5 (+15.4)} \\
        \midrule
        TCL                       & 78.0 & 83.0 & 80.4 & 84.4 \\
        \rowcolor[HTML]{F4CCCC}
        + TTD                     & 80.7 & 82.1 & \textbf{81.4} & \textbf{85.5 (+1.1)} \\
        \bottomrule
      \end{tabular}%
    }%
  \end{subtable}
  \hfill
  \begin{subtable}[t]{0.44\linewidth}
    \centering
    \caption{Text-level Semantic Segmentation.}
    \label{tab:rebuttal_masks}
    \resizebox{\linewidth}{!}{%
      \begin{tabular}{@{}lccc@{}}
        \toprule
        Method & CaptionIoU & mFPR & mFNR \\
        \midrule
        \midrule
        MaskCLIP                  & 59.1 & 0.161 & 0.248 \\
        \rowcolor[HTML]{F4CCCC}
        + TTD                     & \textbf{62.9 (+3.8)} & 0.172 & 0.199 \\
        \midrule
        TCL                       & 71.0 & 0.147 & 0.144 \\
        \rowcolor[HTML]{F4CCCC}
        + TTD                     & \textbf{72.2 (+1.2)} & 0.133 & 0.144 \\
        \bottomrule
      \end{tabular}%
    }%
  \end{subtable}
\end{table}

In this paper, we introduce TTD as a novel method capable of automatically labeling tag- and pixel-level annotations for image-text pairs from the training set, without the need for external pre-trained models beyond the baseline (CLIP), as illustrated in \cref{tab:supp_rel}. Our approach autonomously selects image-relevant tags from the text (\cref{fig:alignment_ann}(a)) and generates a corresponding text mask (\cref{fig:alignment_ann}(c)), highlighting the image areas associated with the text based on tag masks (\cref{fig:alignment_ann}(b)). By eschewing external NLP or vision models in this process, we avoid biases stemming from their training.

Evaluation in \cref{tab:train_labeling} demonstrates the superior quality of tag/mask labels generated by TTD for the train set (CC3M and CC12M datasets). By mitigating single tag bias, our method outperforms previous approaches significantly. Our approach's significance lies in its capacity to automatically generate rich annotations (tags/masks) across various training datasets (\emph{e.g.}, image captioning, medical), serving as a cost-efficient and precise auto-labeler, which is highly beneficial for CLIP-based applications. Thus, it facilitates learning pixel-tag alignment through these automatically generated labels.

\section{Additional Analysis on Single Tag Bias}

\begin{figure}[!t]
  \centering
  \includegraphics[width=\linewidth]{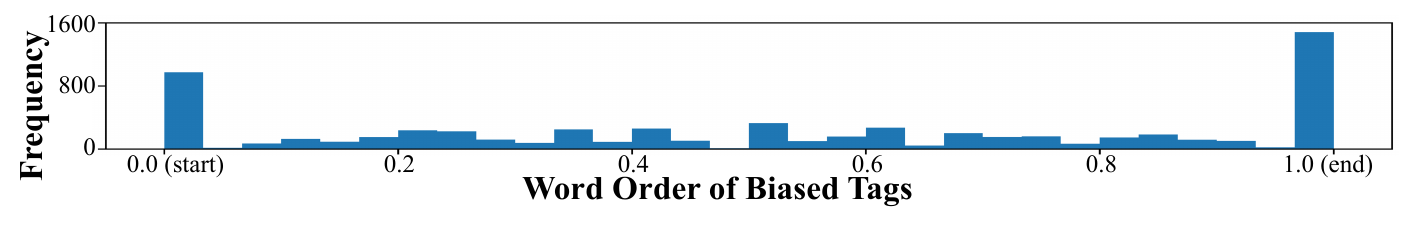}
  \caption{\textbf{Distribution of Order for Word with Single Tag Bias}. Word order within each sentence is normalized from 0 to 1. (\emph{e.g.}, the second word in a three-word caption has a normalized order of 0.5.)
  }
  \label{fig:tag_position}
\end{figure}

We first identify the single tag bias, occurring in nearly $82\%$ of image-text pairs in the CC3M validation dataset (our proposed dataset). This frequency is determined using $s_{\mathrm{image}}$ (\cref{eq:s_image}), where a single tag is selected despite the existence of multiple ground-truth tags, based on non-parametric thresholding introduced in \cref{subsec:tagging}.

Various factors, such as model overfitting and characteristics of the training data, may contribute to single tag bias. It might seem that bias occurs towards words describing elements deemed important in images (\emph{e.g.}, main object), often placed at the beginning of sentences in human-generated descriptions. However, an analysis of the biased tag positions in \cref{fig:tag_position} indicates a scattered distribution throughout sentences, lacking significant correlation with specific positions. Investigating these underlying causes could be an intriguing avenue for future research.

\newpage
\section{Experimental Details}
\label{sec:supp_exp_detail}

\subsection{Implementation Details}
\label{subsec:supp_impl}
\setlength{\tabcolsep}{1.5pt}
\begin{table}[!tb]
  \caption{\textbf{Implementation Details}.}
  \label{tab:supp_impl_details}
  \centering
  \resizebox{\textwidth}{!}{
  \begin{tabular}{>{\centering}p{0.3\textwidth}>{\centering}p{0.35\textwidth}>{\centering\arraybackslash}p{0.35\textwidth}}
    \toprule
     & MaskCLIP~\cite{zhou2022extract} & TCL~\cite{cha2023learning} \\
    \midrule
    \midrule
    Image size & \multicolumn{2}{c}{448 $\times$ 448} \\
    Batch size & \multicolumn{2}{c}{32} \\
    Total epochs & \multicolumn{2}{c}{1} \\
    Optimizer & \multicolumn{2}{c}{AdamW} \\
    Learning rate & \multicolumn{2}{c}{1e-4} \\
    Weight decay & \multicolumn{2}{c}{5e-5} \\
    Data augmentation & \multicolumn{2}{c}{random resize (320-640) $\rightarrow$ random crop (448 $\times$ 448)} \\
    \# params (M) & 150.2 (original: 149.6) & 179.3 (original: 178.3) \\
    Mask post-processing & CRF & PAMR \\
    Prompt template & \multicolumn{2}{c}{no template used during train \& OpenAI template during inference} \\
    \bottomrule
  \end{tabular}
  }
\end{table}


\setlength{\tabcolsep}{5pt}
\begin{table}[!tb]
  \caption{\textbf{Effect of Sample Pruning}.
  }
  \label{tab:supp_sample_pruning}
  \centering
  \resizebox{0.7\textwidth}{!}{
  \begin{tabular}{@{}ccccc@{}}
    \toprule
    Sample pruning & \# samples & training time (hrs) & mIoU @VOC\\
    \midrule
    \midrule
    Baseline & - & - & 55.0 \\
    \xmark & 12,795,608 & 247 & 60.5 (+5.5) \\
    \cmark & 1,758,959 & 34 & \textbf{61.1} (+6.1) \\
    \bottomrule
  \end{tabular}
  }
\end{table}
We conduct model training using 4 NVIDIA A100 GPUs, each having 80GB of memory. Implementation details are summarized in \cref{tab:supp_impl_details}. Most settings, including image size, prompt template usage (for both tag and text input), and post-processing, align with the original setup of baseline models~\cite{zhou2022extract, cha2023learning}. Our batch size is set to 32, the maximum feasible size given our available GPU resources. Training data consists of the CC3M and CC12M datasets~\cite{sharma2018conceptual, changpinyo2021conceptual}, with additional sample pruning to remove noisy samples. We calculate cosine similarity between image and text embeddings for all pairs, sampling those with similarity values exceeding the mean plus the standard deviation. Ultimately, 1.8M pairs are selected from a total of 12.8M pairs, as shown in \cref{tab:supp_sample_pruning}. Training time has been reduced approximately sevenfold, with a performance boost of about +0.6 mIoU, underscoring the efficacy of filtering out noisy samples.

\begin{figure}[!t]
    \centering
    \includegraphics[width=0.9\linewidth]{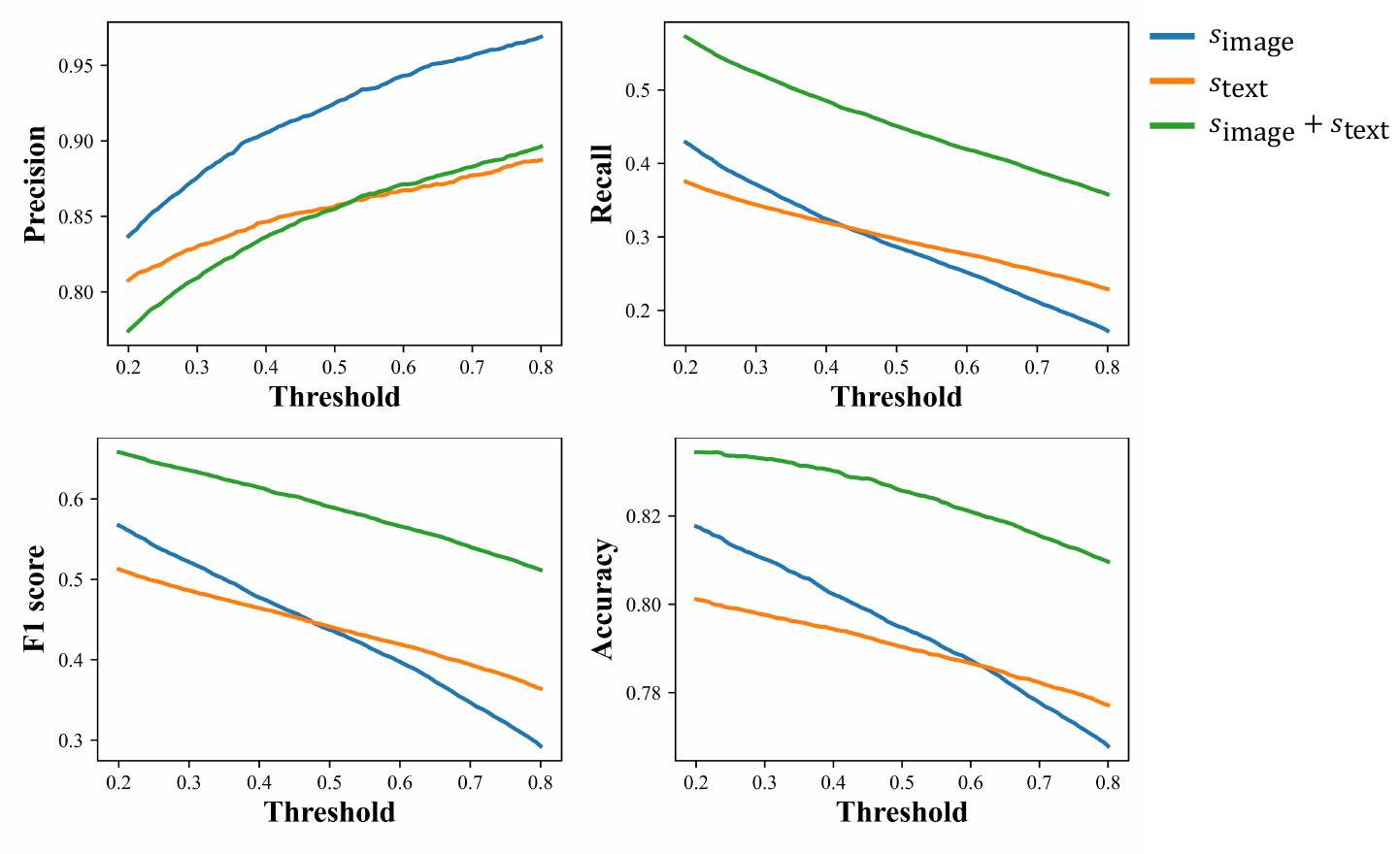}
\caption{\textbf{Performance of Multi-Tag Selection with Varying Thresholds}.}
\label{fig:supp_scoring_thrd}
\end{figure}
\subsection{Evaluation Details}
\label{subsec:supp_eval}
For assessing multi-tag selection (\cref{subsubsec:exp_tagacc}), we utilized two distinct methodologies in the main paper. Firstly, we employed external NLP models such as the NLTK parser~\cite{loper2002nltk} and large language models (LLM)~\cite{chiang2023vicuna, bai2023qwen}. In our utilization of LLM, we adopted identical prompts as utilized in the prior work~\cite{liu2023tagalign}. Secondly, in our scoring-based selection, we set a threshold of 0.5 since the resulting scores ranged from 0 to 1. Our exploration shown in \cref{fig:supp_scoring_thrd} involves adjusting the threshold from 0.2 to 0.8 and observing the corresponding evaluation outcomes. As expected, increasing the threshold led to a reduction in the number of selected tags, resulting in higher precision but lower recall. While fluctuations were observed in F1 score and accuracy, all metrics' optimal values fell short compared to our scoring approach ($s_{\mathrm{pixel}}^{\mathrm{ours}}$ (\cref{eq:s_pixel})) as depicted in \cref{tab:ovc_tagacc}(b). Notably, mAP, being insensitive to the threshold, consistently highlighted the superior performance of our scoring method.

In the evaluation of text-level semantic segmentation (\cref{subsubsec:exp_captioniou}), we maintained a threshold of $0.4$ for TCL-based experiments and $0.5$ for MaskCLIP-based experiments, aligning with their original configurations where regions with similarity values surpassing the threshold are identified as foreground areas.

\newpage
\section{Additional Experiments}
\label{sec:supp_add_exps}
In all experiments presented in this section, we assess the performance of our method (TTD) with the baseline TCL~\cite{cha2023learning}.

\setlength{\tabcolsep}{5pt}
\begin{table}[!tb]
  \caption{\textbf{Effect of Tagging Methods}.
  }
  \label{tab:supp_ab_tagging}
  \centering
  \resizebox{0.7\textwidth}{!}{
  \begin{tabular}{@{}lccccc@{}}
    \toprule
    \multirow{2}{*}{Method} & \multicolumn{3}{c}{Multi-tag selection} & \multicolumn{2}{c}{Segmentation} \\
    \cline{2-4} \cline{5-6}
    & P & R & F1 & CaptionIoU & mIoU @VOC\\
    \midrule
    \midrule
    Baseline~\cite{cha2023learning} & - & - & - & 60.4 & 55.0 \\
    \midrule
    NLTK~\cite{loper2002nltk} & 59.8 & 83.7 & 69.8 & 61.8 & 56.5 \\
    Vicuna-33B~\cite{chiang2023vicuna} & 52.7 & 70.7 & 60.4 & 60.4 & 55.9 \\
    Qwen-72B~\cite{bai2023qwen} & 69.3 & 56.2 & 62.1 & 60.7 & 56.2 \\
    $s_{\mathrm{image}}$ (\cref{eq:s_image}) & 92.5 & 28.6 & 43.7 & 56.3 & 52.5 \\
    $s_{\mathrm{pixel}}^{\mathrm{ours}}$ (\cref{eq:s_pixel}) & 82.9 & 74.5 & \textbf{78.5 }& \textbf{65.5} &  \textbf{61.1} \\
    \bottomrule
  \end{tabular}
  }
\end{table}
\subsection{Effect of Tagging Methods}
\label{subsec:supp_ab_tagging}
As an extension of the experiments outlined in \cref{subsubsec:exp_ab_tagging}, we further investigate the impact of different tagging methods on the performance of the baseline model (TCL) trained with our method. As depicted in \cref{tab:supp_ab_tagging}, there exists a proportional relationship between the effectiveness of tag selection and the resulting CaptionIoU performance. Improved tag selection performance leads to the generation of a more accurate ideal image-text similarity map, effectively alleviating the single tag bias. Furthermore, enhancements in tag-level segmentation performance signify an improved image-text alignment. This observation underscores the critical role of tag selection methods in enhancing the overall performance and alignment of models trained with our approach.

\label{subsec:supp_tag}
\begin{figure}[!t]
    \centering
    \includegraphics[width=\linewidth]{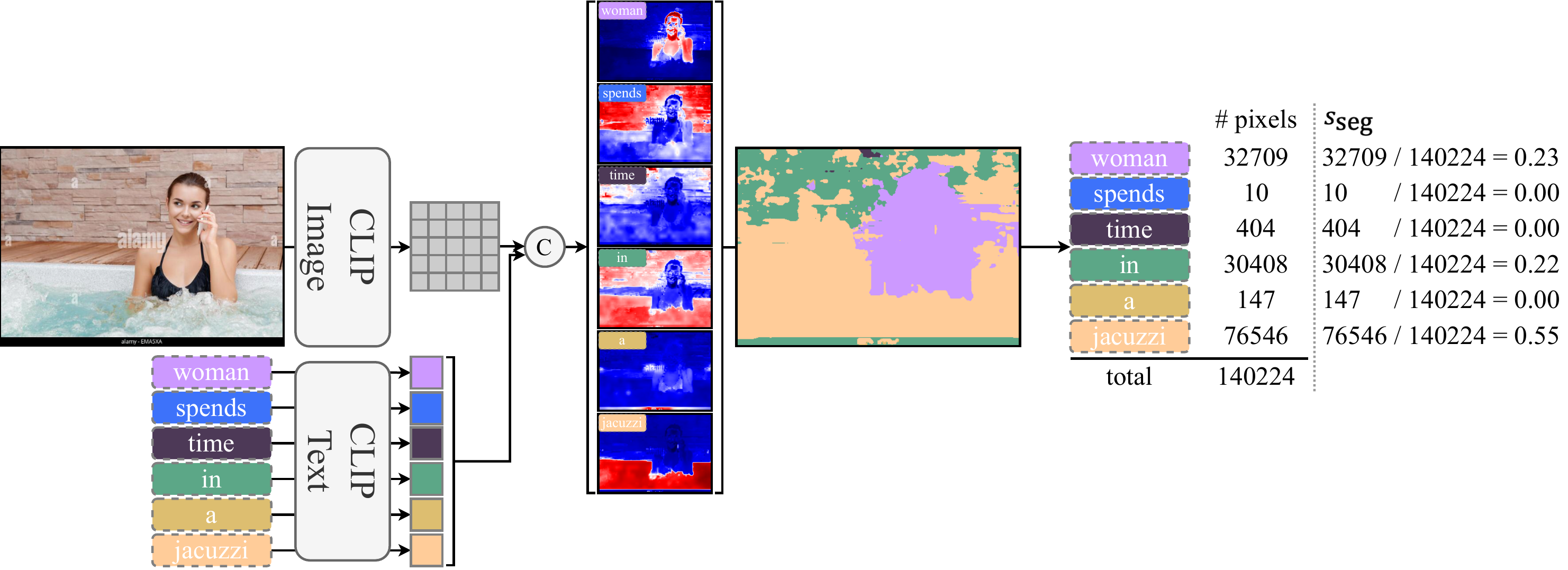}
\caption{\textbf{Illustration of $s_{\mathrm{seg}}$}.}
\label{fig:supp_s_seg}
\end{figure}
\definecolor{pink}{rgb}{0.96,0.82,0.82}

\setlength{\tabcolsep}{1.5pt}
\begin{table}[!tb]
  \caption{\textbf{Multi-Tag Selection with Different Scoring Methods}.}
  \label{tab:supp_tagacc}
  \centering
  \resizebox{0.5\textwidth}{!}{
  \begin{tabular}{@{}lccccc@{}}
    \toprule
    Scoring & P & R & F1 & Acc & mAP\\
    \midrule
    \midrule
    $s_{\mathrm{image}}$ (\cref{eq:s_image})        & 92.5 & 28.6 & 43.7 & 79.5 & 83.2 \\ 
    $s_{\mathrm{text}}$ (\cref{eq:s_text})        & 85.6 & 29.7 & 44.1 & 79.0 & 82.1 \\
    $s_{\mathrm{image}}+s_{\mathrm{text}}$         & 85.5 & 45.1 & 59.0 & 82.6 & 84.5 \\
    $s_{\mathrm{seg}}$         & 48.4 & 11.4 & 18.4 & 71.9 & 57.9 \\
    \midrule
    $s_{\mathrm{pixel}}^{\mathrm{ours}}$ (\cref{eq:s_pixel})        & 82.9 & 74.5 & 78.5 & 88.6 & 90.3 \\
    + TTD (Ours)        & 88.3 & 78.0 & \textbf{82.8} & \textbf{91.0} & \textbf{93.7} \\
  \bottomrule
  \end{tabular}
  }
\end{table}
\subsection{Multi-Tag Selection}
Here, we delve into an additional scoring method and analysis of multi-tag selection not covered in the main paper. We explore the performance of previously introduced scoring techniques ($s_{\mathrm{image}}$ (\cref{eq:s_image}), $s_{\mathrm{text}}$ (\cref{eq:s_text})) across various thresholds in \cref{subsec:supp_eval}.

As mentioned in \cref{sec:intro}, each tag can be scored based on the proportion of image pixels similar to that tag ($s_{\mathrm{seg}}$). Specifically, as depicted in \cref{fig:supp_s_seg}, we generate a similarity map between the image and each tag. Subsequently, for every pixel, we identify the most relevant tag with the highest similarity value, enabling us to score each tag based on the proportion of its relevant pixels. As illustrated in \cref{tab:supp_ab_tagging}, $s_{\mathrm{seg}}$ demonstrates relatively high precision since it accentuates a single tag. However, due to the tendency of irrelevant tags to occupy more pixels than relevant ones, it encounters challenges in score ranking, resulting in a significantly low mAP (\emph{i.e.}, 57.9). In contrast, our scoring method ($s_{\mathrm{pixel}}$) effectively mitigates the single tag bias while maintaining correct ranking, resulting in a high F1 score and mAP.

\begin{figure}[t]
  \centering
  \includegraphics[width=\linewidth]{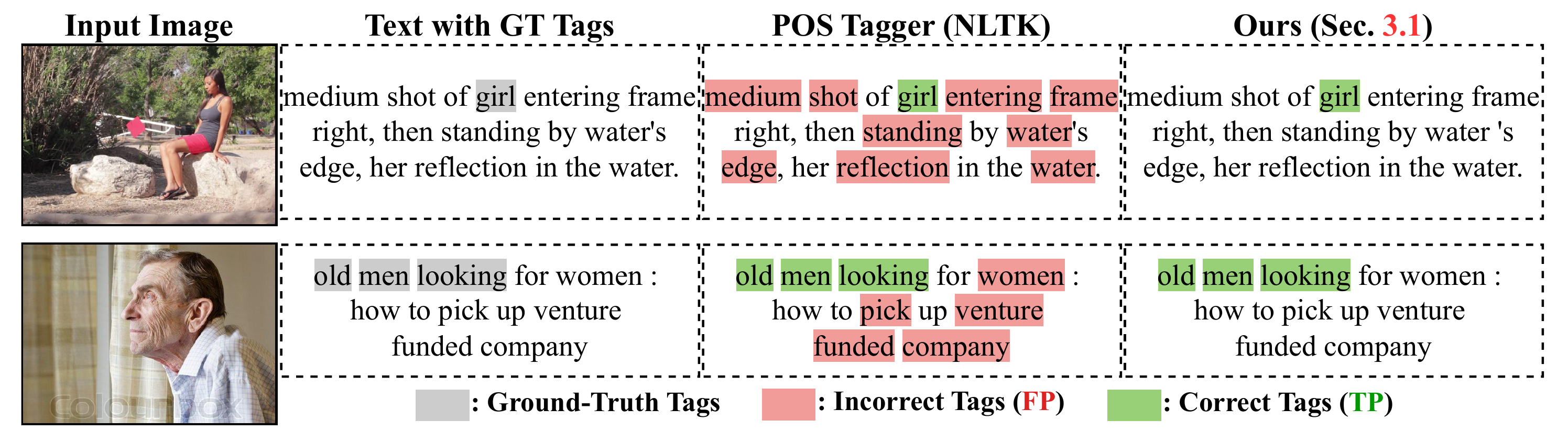}
  \caption{\textbf{Comparison with POS tagger and our TTD}.}
  \label{fig:pos_prob}
\end{figure}

\setlength{\tabcolsep}{1.5pt}
\begin{table}[t]
  \caption{\textbf{Multi-Tag Selection Performance}.}
  \label{tab:templates}
  \centering
  \resizebox{\linewidth}{!}{
    \begin{tabular}{@{}lccccc@{}}
      \toprule
      Method & Encoding tags with templates & P & R & F1 & Acc \\
      \midrule
      \midrule
      (a) NLTK \tiny{(semantically meaningful tags)}      & -        & 43.5 & 99.5 & 60.6 & 64.5 \\
      (b) TCL+TTD   & \xmark   & 87.3 & 75.4 & 80.9 & 90.1 \\
      (c) TCL+TTD   & \cmark   & 88.3 & 78.0 & \textbf{82.8 (+1.9)} & \textbf{ 91.0 (+0.9)} \\
      \bottomrule
    \end{tabular}%
  }
\end{table}

Existing tag selection methods rely solely on NLP modules to extract nouns from the given text. While it may seem intuitive to extract all semantically meaningful tags beyond nouns, many of them may not be relevant to the image content, as shown in \cref{fig:pos_prob} and \cref{tab:templates}(a). For example, even in cases where a woman is absent from the image, the POS tagger still identifies it as an image-related tag. (second image in \cref{fig:pos_prob}) In contrast, our TTD accurately extracts image-relevant tags, demonstrating its superiority over the simple extraction methods using POS tagger (\cref{tab:templates}(c)).

Furthermore, we examine the influence of text templates on encoding tags and text inputs. Our findings in \cref{tab:templates} demonstrate that template usage improves the performance of multi-tag selection. This aligns with the observation that CLIP typically excels at encoding complete sentences or descriptions rather than individual tags.


\setlength{\tabcolsep}{1.5pt}
\begin{table}[tb]
  \caption{\textbf{Open-Vocabulary Multi-Label Classification on NUS-WIDE~\cite{nus-wide-civr09}}. $*$ indicates the results based on our reproduction. Ann., use of a pre-defined label set during training.}
  \label{tab:supp_ovc_multi}
  \centering
  \resizebox{0.5\textwidth}{!}{
    \begin{tabular}{@{}lccc@{}}
      \toprule
      Method & Ann. & Backbone & mAP \\
      \midrule
      \midrule
      ADDS~\cite{xu2022dual}          & \cmark & ViT-B/32 & 36.56 \\
      MKT~\cite{he2023open}           & \cmark & ViT-B/16 & 37.6 \\
      DualCoOp~\cite{sun2022dualcoop}           & \cmark & ResNet-50 & 43.6 \\
      TaI-DPT~\cite{guo2023texts}           & \cmark & ResNet-50 & 44.99 \\
      \midrule
      CLIP~\cite{radford2021learning}  & \xmark & ResNet-50 & 35.58 \\
      CLIP Surgery~\cite{li2023clip}  & \xmark & ResNet-50 & 39.99 \\
      CLIP Surgery$^{*}$~\cite{li2023clip}  & \xmark & ViT-B/16 & 40.53 \\
      \textbf{Ours} (TCL~\cite{cha2023learning}+TTD)          & \xmark & ViT-B/16 & \textbf{42.63} \\
      \bottomrule
    \end{tabular}%
  }
\end{table}
\subsection{Open-Vocabulary Multi-label Classification}
\label{subsec:supp_ovc_multi}
We contrast our method with existing CLIP-based multi-label classification approaches in \cref{tab:supp_ovc_multi}. Notably, all models except CLIP and CLIP Surgery~\cite{li2023clip} rely on a predefined label set during training, requiring subsequent fine-tuning on seen labels. To ensure fairness in comparison, we reproduce the result of CLIP Surgery without using the label set during inference, aligning with our approach's methodology.\footnote{The original approach of CLIP Surgery requires feature correction using a predefined label set, so we use feature correction with an empty sentence instead.} Leveraging the same backbone (ViT-B/16), our method surpasses both CLIP and CLIP Surgery, achieving an mAP of 42.63. This performance superiority highlights the effectiveness and efficiency of our approach in multi-label classification tasks, without the need for additional fine-tuning or label annotations. Our method not only outperforms the existing techniques but also demonstrates its capacity to address classification challenges with minimal data dependencies.

\definecolor{pink}{rgb}{0.96,0.82,0.82}

\setlength{\tabcolsep}{1.5pt}
\begin{table}[!tb]
  \caption{\textbf{Open-Vocabulary Semantic Segmentation on Referring Datasets}. 
  We use ViT-B/16 as the backbone for all methods. 
  }
  \label{tab:supp_ovs_refer}
  \centering
  \resizebox{0.9\textwidth}{!}{
  \begin{tabular}{@{}lcccccccc@{}}
    \toprule
    \multirow{2}{*}{Method} & \multicolumn{3}{c}{RefCOCO} & \multicolumn{3}{c}{RefCOCO+} & RefCOCOg & \multirow{2}{*}{Average} \\
    \cline{2-4} \cline{5-7} \cline{8-8}
     & val & testA & testB & val & testA & testB & val & \\
    \midrule
    \midrule
    GroupViT~\cite{xu2022groupvit}      &  7.99 &  6.16 & 10.51 &  8.49 &  6.79 & 10.59 & 10.68 & 8.74 \\
    MaskCLIP~\cite{radford2021learning} & 11.52 & 11.85 & 12.06 & 11.87 & 12.01 & 12.57 & 12.74 & 12.09 \\
    TagAlign~\cite{liu2023tagalign}     & \textbf{18.75} & \textbf{20.31} & \underline{20.64} & \underline{19.24} & \underline{20.88} & \underline{21.23} & \underline{23.69} & \underline{20.68} \\
    \midrule
    \textbf{Ours} (TCL~\cite{cha2023learning}+TTD) & \underline{18.64} & \underline{13.17} & \textbf{25.96} & \textbf{19.75} & \underline{14.86} & \textbf{26.35} & \textbf{28.22} & \textbf{20.99} \\
  \bottomrule
  \end{tabular}
  }
\end{table}
\subsection{Open-Vocabulary Semantic Segmentation on Referring Datasets}
\label{subsec:supp_ovs}
We evaluate our model, which is fine-tuned TCL~\cite{cha2023learning} with our method (TTD), on referring datasets~\cite{mao2016generation, yu2016modeling}. Unlike the text-level and tag-level segmentations, referring segmentation involves delineating entities described in free-form text. This task demands a deeper understanding of the textual context to accurately segment the corresponding entities in the image. As depicted in \cref{tab:supp_ovs_refer}, our model surpasses existing approaches, showcasing its proficiency in comprehending textual descriptions and effectively translating them into precise image segmentations. This demonstrates the robustness and versatility of our method across diverse segmentation tasks, further validating its potential for real-world applications where understanding complex textual descriptions is paramount.


\newpage
\begin{figure}[!t]
    \centering
    \includegraphics[width=0.8\linewidth]{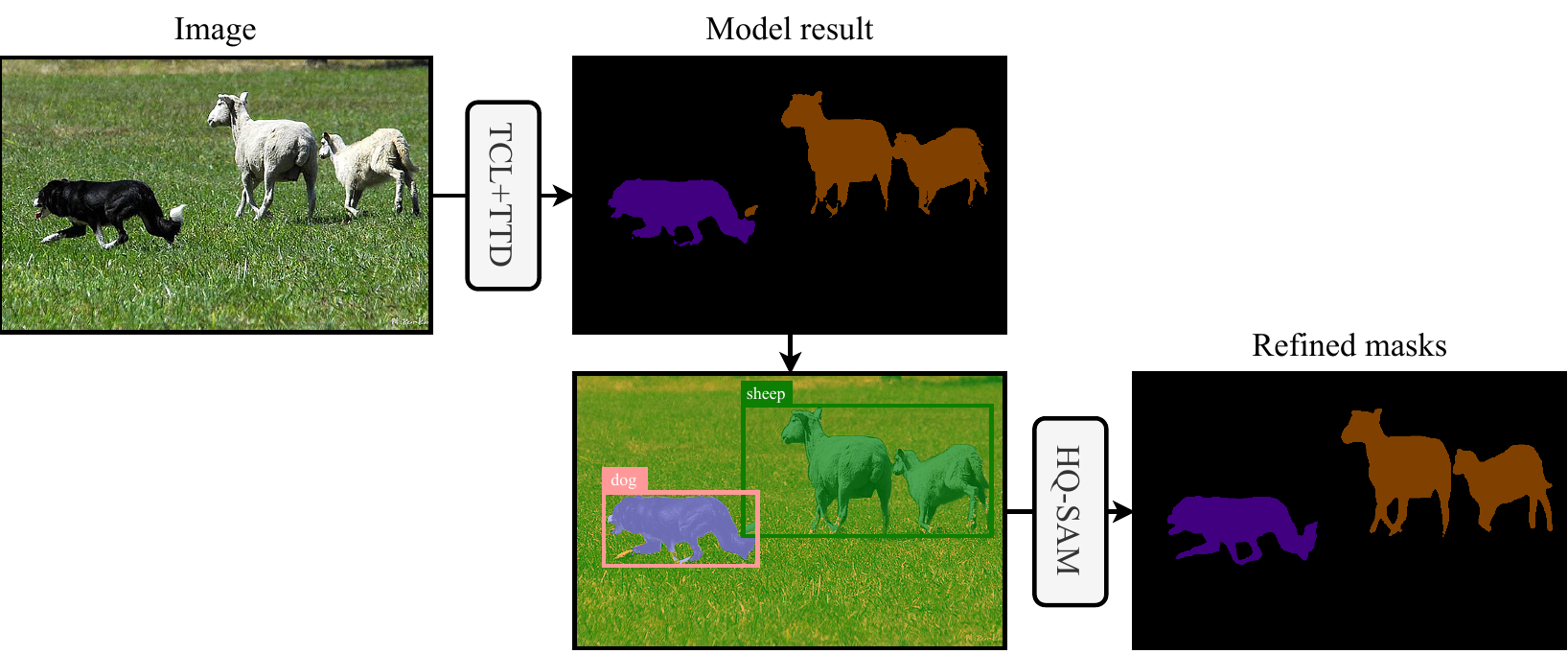}
\caption{\textbf{Refinement Process using HQ-SAM~\cite{ke2024segment}}. Based on the mask outputs of our model, we generate corresponding bounding boxes for each mask. Then, we generate refined masks using HQ-SAM with masks and bounding boxes. }
\label{fig:supp_sam_process}
\end{figure}
\setlength{\tabcolsep}{1.5pt}
\begin{table}[tb]
  \caption{\textbf{Effect of Additional Refinement using HQ-SAM}.}
  \label{tab:supp_sam}
  \centering
  \resizebox{0.6\linewidth}{!}{%
  \begin{tabular}{@{}lcccc@{}}
    \toprule
    Baseline & TTD (Ours) & HQ-SAM~\cite{ke2024segment} & mIoU @VOC\\
    \midrule
    \midrule
    \multirow{3}{*}{MaskCLIP~\cite{zhou2022extract}} & \xmark & \xmark & 29.3 \\
    & \cmark & \xmark & 43.1 \\
    & \cmark & \cmark & 53.7 \\
    \midrule
    \multirow{3}{*}{TCL~\cite{cha2023learning}} & \xmark & \xmark & 55.0 \\
    & \cmark & \xmark & 61.1 \\
    & \cmark & \cmark & 63.4 \\
    \bottomrule
  \end{tabular}%
  }
\end{table}
\subsection{Additional Refinement with SAM}
\label{subsec:supp_sam}
\begin{figure}[!t]
    \centering
    \includegraphics[width=0.9\linewidth]{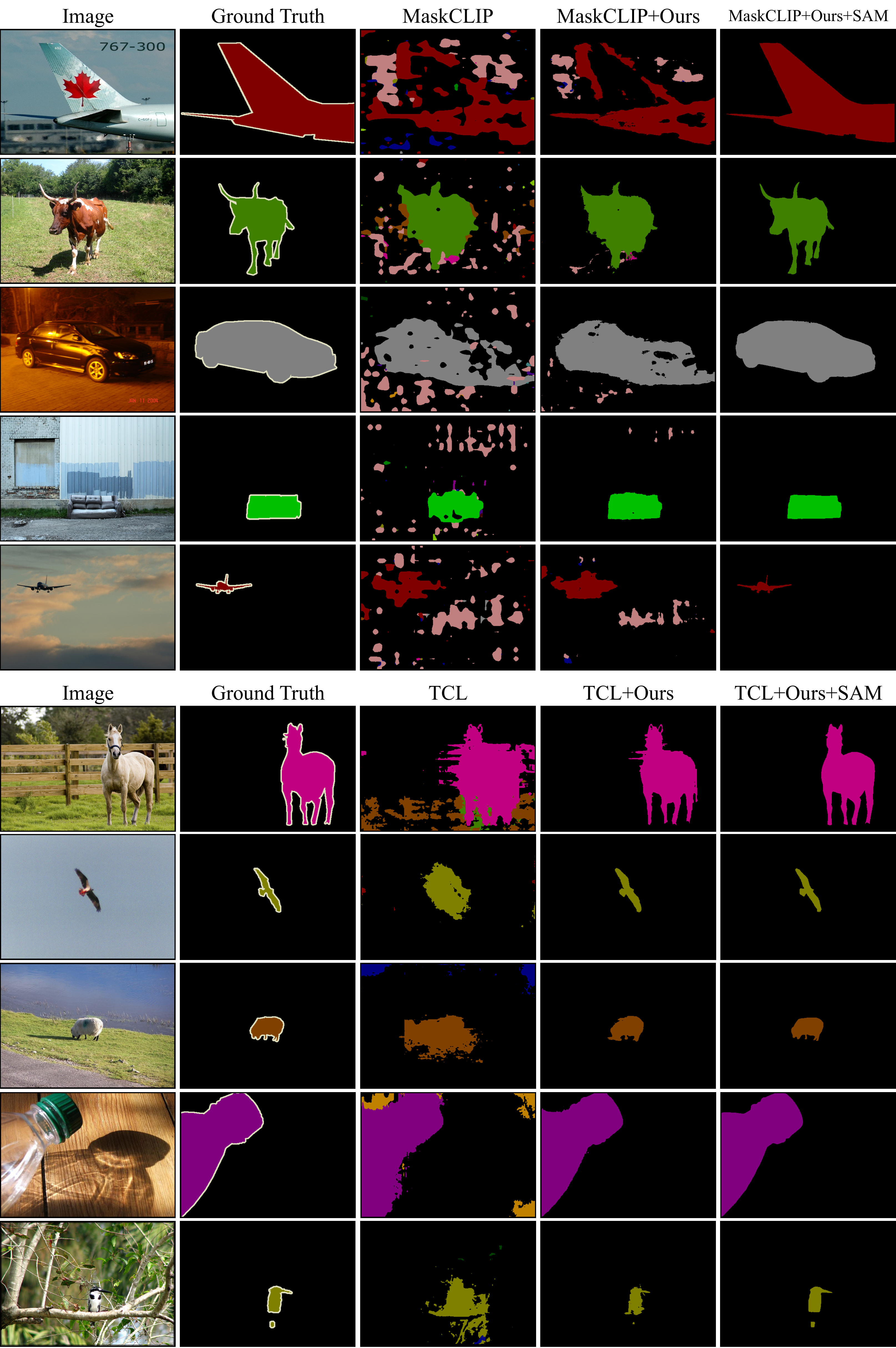}
\caption{\textbf{Additional Qualitative Results on Semantic Segmentation on Pascal VOC, when refined with HQ-SAM~\cite{ke2024segment}}.}
\label{fig:supp_qual_sam}
\end{figure}

We further enhance the segmentation outputs of our model by refining them using HQ-SAM~\cite{ke2024segment} as shown in \cref{fig:supp_sam_process}. This involves feeding the mask output from our model and its corresponding bounding box into HQ-SAM to generate a refined mask. As illustrated in \cref{fig:supp_qual_sam}, the refined masks exhibit sharper object boundaries compared to those produced solely by our method. Given that the MaskCLIP-based model initially performs poorer than the TCL-based model, the refinement's impact is more pronounced, where noisy masks are also removed. Quantitative results in \cref{tab:supp_sam} further support this observation, showing a significant increase of 10.6 mIoU for the MaskCLIP-based model and 2.3 for the TCL-based model. This refinement process with HQ-SAM contributes to cleaner and more accurate segmentation results, enhancing the overall performance of our approach.

\newpage

\section{Generating Tag and Mask Annotations}
\label{sec:supp_dataset}
\begin{figure}[!t]
    \centering
    \includegraphics[width=0.8\linewidth]{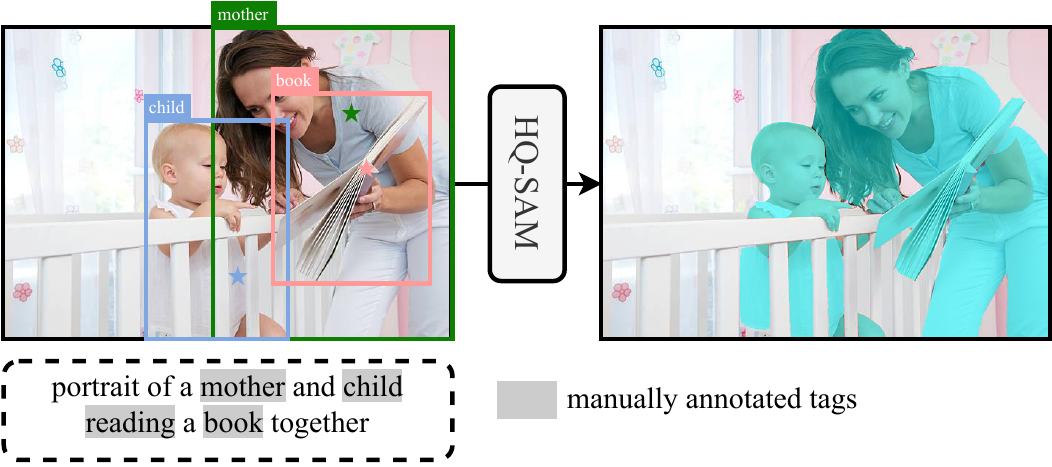}
\caption{\textbf{Generation of Mask Annotations using HQ-SAM}. Beginning with the manually annotated tags—mother, child, reading, and book, as depicted in the figure—we manually outline the bounding boxes and points corresponding to each tag. Subsequently, by merging the masks generated from HQ-SAM, we obtain the final mask representing the relationship between the image and the text.}
\label{fig:supp_benchmark}
\end{figure}

For the CC3M validation dataset~\cite{sharma2018conceptual}, we meticulously annotated tags (words) reflecting the relationship between an image and its corresponding text. Subsequently, we manually delineated bounding boxes and points corresponding to each tag, as illustrated in \cref{fig:supp_benchmark}. Using HQ-SAM~\cite{ke2024segment}, we generated masks for each tag, with the union of all tag masks forming the mask illustrating the relationship between the image and the text. Finally, we filtered out 680 masks of high quality.

Our benchmark stands as the first to include tag annotation for each text along with its mask annotation. Moreover, to our knowledge, no other benchmark addresses segmentation with \textit{text} involving \textit{multiple objects}, crucial for assessing the model's representation of image-text alignment.

\newpage
\section{Additional Qualitative Results}
\label{sec:supp_qual}

\begin{figure}[!ht]
    \centering
    \includegraphics[width=\linewidth]{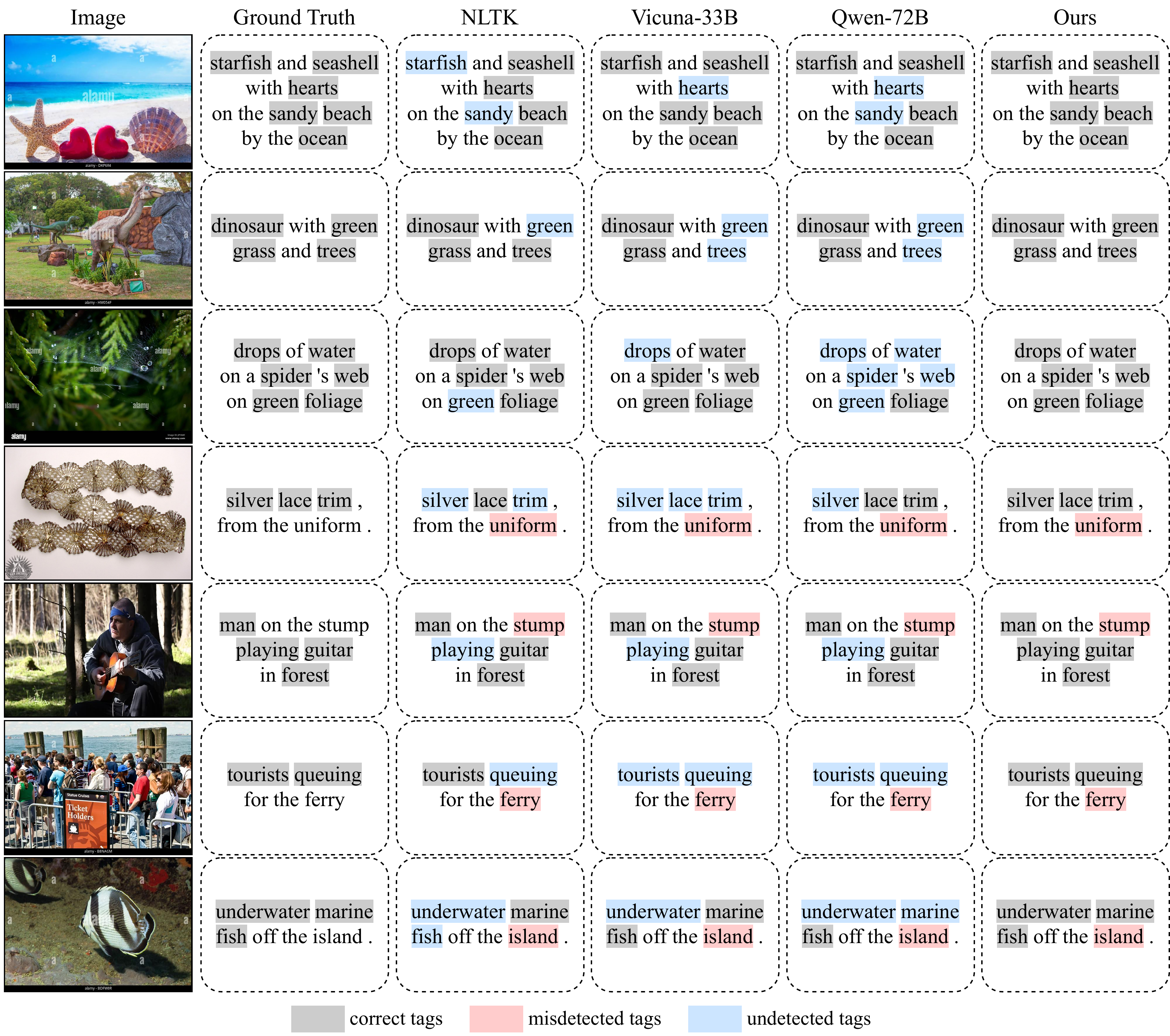}
\caption{\textbf{Additional Qualitative Results on Multi-Tag Selection}.}
\label{fig:supp_qual_tag}
\end{figure}

\begin{figure}[!ht]
    \centering
    \includegraphics[width=\linewidth]{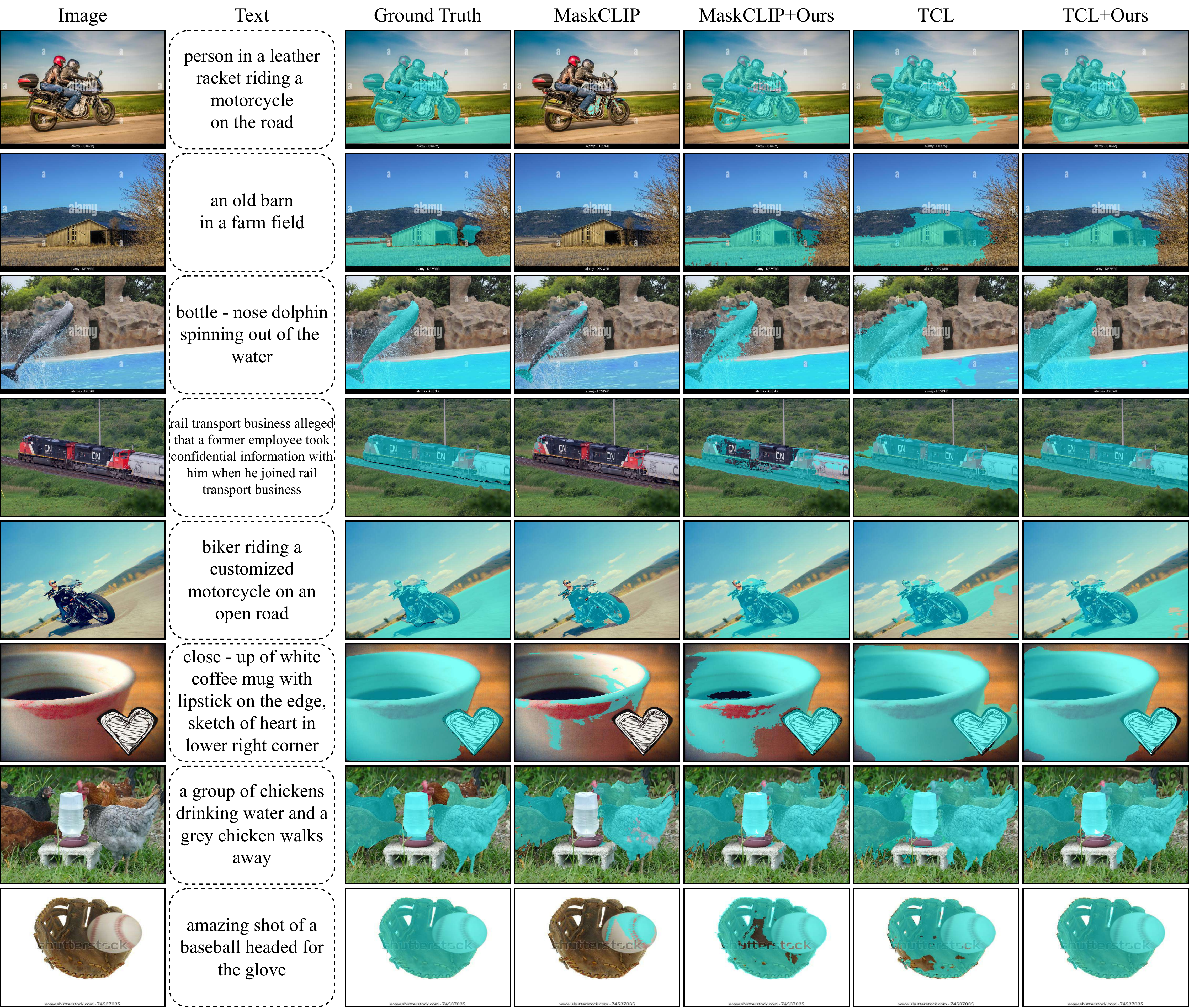}
\caption{\textbf{Additional Qualitative Results on Text-level Segmentation}.}
\label{fig:supp_qual_textseg}
\end{figure}

\clearpage  
\bibliographystyle{splncs04}
\bibliography{main}

\begin{thebibliography}{10}
\providecommand{\url}[1]{\texttt{#1}}
\providecommand{\urlprefix}{URL }
\providecommand{\doi}[1]{https://doi.org/#1}

\bibitem{akbik2019flair}
Akbik, A., Bergmann, T., Blythe, D., Rasul, K., Schweter, S., Vollgraf, R.: {FLAIR}: An easy-to-use framework for state-of-the-art {NLP}. In: {NAACL} 2019, 2019 Annual Conference of the North American Chapter of the Association for Computational Linguistics (Demonstrations). pp. 54--59 (2019)

\bibitem{bai2023qwen}
Bai, J., Bai, S., Chu, Y., Cui, Z., Dang, K., Deng, X., Fan, Y., Ge, W., Han, Y., Huang, F., et~al.: Qwen technical report. arXiv preprint arXiv:2309.16609  (2023)

\bibitem{caesar2018coco}
Caesar, H., Uijlings, J., Ferrari, V.: Coco-stuff: Thing and stuff classes in context. In: Proceedings of the IEEE conference on computer vision and pattern recognition. pp. 1209--1218 (2018)

\bibitem{cha2023learning}
Cha, J., Mun, J., Roh, B.: Learning to generate text-grounded mask for open-world semantic segmentation from only image-text pairs. In: Proceedings of the IEEE/CVF Conference on Computer Vision and Pattern Recognition. pp. 11165--11174 (2023)

\bibitem{changpinyo2021conceptual}
Changpinyo, S., Sharma, P., Ding, N., Soricut, R.: Conceptual 12m: Pushing web-scale image-text pre-training to recognize long-tail visual concepts. In: Proceedings of the IEEE/CVF Conference on Computer Vision and Pattern Recognition. pp. 3558--3568 (2021)

\bibitem{chen2023exploring}
Chen, J., Zhu, D., Qian, G., Ghanem, B., Yan, Z., Zhu, C., Xiao, F., Culatana, S.C., Elhoseiny, M.: Exploring open-vocabulary semantic segmentation from clip vision encoder distillation only. In: Proceedings of the IEEE/CVF International Conference on Computer Vision. pp. 699--710 (2023)

\bibitem{chiang2023vicuna}
Chiang, W.L., Li, Z., Lin, Z., Sheng, Y., Wu, Z., Zhang, H., Zheng, L., Zhuang, S., Zhuang, Y., Gonzalez, J.E., et~al.: Vicuna: An open-source chatbot impressing gpt-4 with 90\%* chatgpt quality. See https://vicuna. lmsys. org (accessed 14 April 2023)  (2023)

\bibitem{nus-wide-civr09}
Chua, T.S., Tang, J., Hong, R., Li, H., Luo, Z., Zheng, Y.T.: Nus-wide: A real-world web image database from national university of singapore. In: Proc. of ACM Conf. on Image and Video Retrieval (CIVR'09). Santorini, Greece. (July 8-10, 2009)

\bibitem{conti2024vocabulary}
Conti, A., Fini, E., Mancini, M., Rota, P., Wang, Y., Ricci, E.: Vocabulary-free image classification. Advances in Neural Information Processing Systems  \textbf{36} (2024)

\bibitem{cordts2016cityscapes}
Cordts, M., Omran, M., Ramos, S., Rehfeld, T., Enzweiler, M., Benenson, R., Franke, U., Roth, S., Schiele, B.: The cityscapes dataset for semantic urban scene understanding. In: Proceedings of the IEEE conference on computer vision and pattern recognition. pp. 3213--3223 (2016)

\bibitem{ding2022decoupling}
Ding, J., Xue, N., Xia, G.S., Dai, D.: Decoupling zero-shot semantic segmentation. In: Proceedings of the IEEE/CVF Conference on Computer Vision and Pattern Recognition. pp. 11583--11592 (2022)

\bibitem{everingham2010pascal}
Everingham, M., Van~Gool, L., Williams, C.K., Winn, J., Zisserman, A.: The pascal visual object classes (voc) challenge. International journal of computer vision  \textbf{88},  303--338 (2010)

\bibitem{guo2023mvp}
Guo, J., Wang, Q., Gao, Y., Jiang, X., Tang, X., Hu, Y., Zhang, B.: Mvp-seg: Multi-view prompt learning for open-vocabulary semantic segmentation. arXiv preprint arXiv:2304.06957  (2023)

\bibitem{guo2023texts}
Guo, Z., Dong, B., Ji, Z., Bai, J., Guo, Y., Zuo, W.: Texts as images in prompt tuning for multi-label image recognition. In: Proceedings of the IEEE/CVF Conference on Computer Vision and Pattern Recognition. pp. 2808--2817 (2023)

\bibitem{he2023open}
He, S., Guo, T., Dai, T., Qiao, R., Shu, X., Ren, B., Xia, S.T.: Open-vocabulary multi-label classification via multi-modal knowledge transfer. In: Proceedings of the AAAI Conference on Artificial Intelligence. vol.~37, pp. 808--816 (2023)

\bibitem{spacy2}
Honnibal, M., Montani, I.: {spaCy 2}: Natural language understanding with {B}loom embeddings, convolutional neural networks and incremental parsing (2017), to appear

\bibitem{hu2021lora}
Hu, E.J., Shen, Y., Wallis, P., Allen-Zhu, Z., Li, Y., Wang, S., Wang, L., Chen, W.: Lora: Low-rank adaptation of large language models. arXiv preprint arXiv:2106.09685  (2021)

\bibitem{ke2024segment}
Ke, L., Ye, M., Danelljan, M., Tai, Y.W., Tang, C.K., Yu, F., et~al.: Segment anything in high quality. Advances in Neural Information Processing Systems  \textbf{36} (2024)

\bibitem{kirillov2023segment}
Kirillov, A., Mintun, E., Ravi, N., Mao, H., Rolland, C., Gustafson, L., Xiao, T., Whitehead, S., Berg, A.C., Lo, W.Y., et~al.: Segment anything. arXiv preprint arXiv:2304.02643  (2023)

\bibitem{kobayashi2023two}
Kobayashi, T.: Two-way multi-label loss. In: Proceedings of the IEEE/CVF Conference on Computer Vision and Pattern Recognition. pp. 7476--7485 (2023)

\bibitem{li2023patchct}
Li, M., Wang, D., Liu, X., Zeng, Z., Lu, R., Chen, B., Zhou, M.: Patchct: Aligning patch set and label set with conditional transport for multi-label image classification. In: Proceedings of the IEEE/CVF International Conference on Computer Vision. pp. 15348--15358 (2023)

\bibitem{li2023clip}
Li, Y., Wang, H., Duan, Y., Li, X.: Clip surgery for better explainability with enhancement in open-vocabulary tasks. arXiv preprint arXiv:2304.05653  (2023)

\bibitem{liang2023open}
Liang, F., Wu, B., Dai, X., Li, K., Zhao, Y., Zhang, H., Zhang, P., Vajda, P., Marculescu, D.: Open-vocabulary semantic segmentation with mask-adapted clip. In: Proceedings of the IEEE/CVF Conference on Computer Vision and Pattern Recognition. pp. 7061--7070 (2023)

\bibitem{liu2023tagalign}
Liu, Q., Zheng, K., Wei, W., Tong, Z., Liu, Y., Chen, W., Wang, Z., Shen, Y.: Tagalign: Improving vision-language alignment with multi-tag classification. arXiv preprint arXiv:2312.14149  (2023)

\bibitem{liu2022open}
Liu, Q., Wen, Y., Han, J., Xu, C., Xu, H., Liang, X.: Open-world semantic segmentation via contrasting and clustering vision-language embedding. In: European Conference on Computer Vision. pp. 275--292. Springer (2022)

\bibitem{loper2002nltk}
Loper, E., Bird, S.: Nltk: The natural language toolkit. arXiv preprint cs/0205028  (2002)

\bibitem{luo2023segclip}
Luo, H., Bao, J., Wu, Y., He, X., Li, T.: Segclip: Patch aggregation with learnable centers for open-vocabulary semantic segmentation. In: International Conference on Machine Learning. pp. 23033--23044. PMLR (2023)

\bibitem{mao2016generation}
Mao, J., Huang, J., Toshev, A., Camburu, O., Yuille, A.L., Murphy, K.: Generation and comprehension of unambiguous object descriptions. In: Proceedings of the IEEE conference on computer vision and pattern recognition. pp. 11--20 (2016)

\bibitem{mottaghi2014role}
Mottaghi, R., Chen, X., Liu, X., Cho, N.G., Lee, S.W., Fidler, S., Urtasun, R., Yuille, A.: The role of context for object detection and semantic segmentation in the wild. In: Proceedings of the IEEE conference on computer vision and pattern recognition. pp. 891--898 (2014)

\bibitem{mukhoti2023open}
Mukhoti, J., Lin, T.Y., Poursaeed, O., Wang, R., Shah, A., Torr, P.H., Lim, S.N.: Open vocabulary semantic segmentation with patch aligned contrastive learning. In: Proceedings of the IEEE/CVF Conference on Computer Vision and Pattern Recognition. pp. 19413--19423 (2023)

\bibitem{radford2021learning}
Radford, A., Kim, J.W., Hallacy, C., Ramesh, A., Goh, G., Agarwal, S., Sastry, G., Askell, A., Mishkin, P., Clark, J., et~al.: Learning transferable visual models from natural language supervision. In: International conference on machine learning. pp. 8748--8763. PMLR (2021)

\bibitem{ranasinghe2023perceptual}
Ranasinghe, K., McKinzie, B., Ravi, S., Yang, Y., Toshev, A., Shlens, J.: Perceptual grouping in contrastive vision-language models. In: Proceedings of the IEEE/CVF International Conference on Computer Vision. pp. 5571--5584 (2023)

\bibitem{ren2023viewco}
Ren, P., Li, C., Xu, H., Zhu, Y., Wang, G., Liu, J., Chang, X., Liang, X.: Viewco: Discovering text-supervised segmentation masks via multi-view semantic consistency. arXiv preprint arXiv:2302.10307  (2023)

\bibitem{sharma2018conceptual}
Sharma, P., Ding, N., Goodman, S., Soricut, R.: Conceptual captions: A cleaned, hypernymed, image alt-text dataset for automatic image captioning. In: Proceedings of the 56th Annual Meeting of the Association for Computational Linguistics (Volume 1: Long Papers). pp. 2556--2565 (2018)

\bibitem{shin2022reco}
Shin, G., Xie, W., Albanie, S.: Reco: Retrieve and co-segment for zero-shot transfer. Advances in Neural Information Processing Systems  \textbf{35},  33754--33767 (2022)

\bibitem{sun2022dualcoop}
Sun, X., Hu, P., Saenko, K.: Dualcoop: Fast adaptation to multi-label recognition with limited annotations. Advances in Neural Information Processing Systems  \textbf{35},  30569--30582 (2022)

\bibitem{xing2024rewrite}
Xing, Y., Kang, J., Xiao, A., Nie, J., Shao, L., Lu, S.: Rewrite caption semantics: Bridging semantic gaps for language-supervised semantic segmentation. Advances in Neural Information Processing Systems  \textbf{36} (2024)

\bibitem{xu2022groupvit}
Xu, J., De~Mello, S., Liu, S., Byeon, W., Breuel, T., Kautz, J., Wang, X.: Groupvit: Semantic segmentation emerges from text supervision. In: Proceedings of the IEEE/CVF Conference on Computer Vision and Pattern Recognition. pp. 18134--18144 (2022)

\bibitem{xu2023learning}
Xu, J., Hou, J., Zhang, Y., Feng, R., Wang, Y., Qiao, Y., Xie, W.: Learning open-vocabulary semantic segmentation models from natural language supervision. In: Proceedings of the IEEE/CVF Conference on Computer Vision and Pattern Recognition. pp. 2935--2944 (2023)

\bibitem{xu2022dual}
Xu, S., Li, Y., Hsiao, J., Ho, C., Qi, Z.: A dual modality approach for (zero-shot) multi-label classification. arXiv preprint arXiv:2208.09562  (2022)

\bibitem{yi2023simple}
Yi, M., Cui, Q., Wu, H., Yang, C., Yoshie, O., Lu, H.: A simple framework for text-supervised semantic segmentation. In: Proceedings of the IEEE/CVF Conference on Computer Vision and Pattern Recognition. pp. 7071--7080 (2023)

\bibitem{yu2016modeling}
Yu, L., Poirson, P., Yang, S., Berg, A.C., Berg, T.L.: Modeling context in referring expressions. In: Computer Vision--ECCV 2016: 14th European Conference, Amsterdam, The Netherlands, October 11-14, 2016, Proceedings, Part II 14. pp. 69--85. Springer (2016)

\bibitem{zhou2019semantic}
Zhou, B., Zhao, H., Puig, X., Xiao, T., Fidler, S., Barriuso, A., Torralba, A.: Semantic understanding of scenes through the ade20k dataset. International Journal of Computer Vision  \textbf{127},  302--321 (2019)

\bibitem{zhou2022extract}
Zhou, C., Loy, C.C., Dai, B.: Extract free dense labels from clip. In: European Conference on Computer Vision. pp. 696--712. Springer (2022)

\end{thebibliography}

\end{document}